%% file: main.tex
\documentclass[10pt,twocolumn,letterpaper]{article}

\usepackage[pagenumbers]{cvpr} 

\definecolor{cvprblue}{rgb}{0.21,0.49,0.74}
\usepackage[pagebackref,breaklinks,colorlinks,allcolors=cvprblue]{hyperref}
\usepackage{algorithm,algorithmic}
\usepackage{multirow}
\usepackage{multicol}

\def\paperID{12241} 
\def\confName{CVPR}
\def\confYear{2025}

\makeatletter
\def\blfootnote{\gdef\@thefnmark{}\@footnotetext}
\makeatother

\title{StyleStudio: Text-Driven Style Transfer with Selective Control of Style Elements\vspace{-4mm}}

\author{
Mingkun Lei\textsuperscript{\rm 1} \quad Xue Song\textsuperscript{\rm 2} \quad Beier Zhu\textsuperscript{\rm 1, \rm 3} \quad Hao Wang\textsuperscript{\rm 4} \quad Chi Zhang\textsuperscript{\rm 1 *} \\
\textsuperscript{\rm 1} AGI Lab, Westlake University \quad
\textsuperscript{\rm 2} Fudan University \quad
\textsuperscript{\rm 3} Nanyang Technological University \quad \\
\textsuperscript{\rm 4} The Hong Kong University of Science and Technology (Guangzhou) \quad \\
{\tt\small \{leimingkun, chizhang\}@westlake.edu.cn \quad xuesong21@m.fudan.edu.cn} \\
{\tt\small beier.zhu@ntu.edu.sg \quad haowang@hkust-gz.edu.cn} \\
\url{https://stylestudio-official.github.io/}
}

\begin{document}
\twocolumn[{%
\renewcommand\twocolumn[1][]{#1}%
\maketitle
\vspace{-13mm}
\begin{center}
    \centering
    \includegraphics[width=\linewidth]{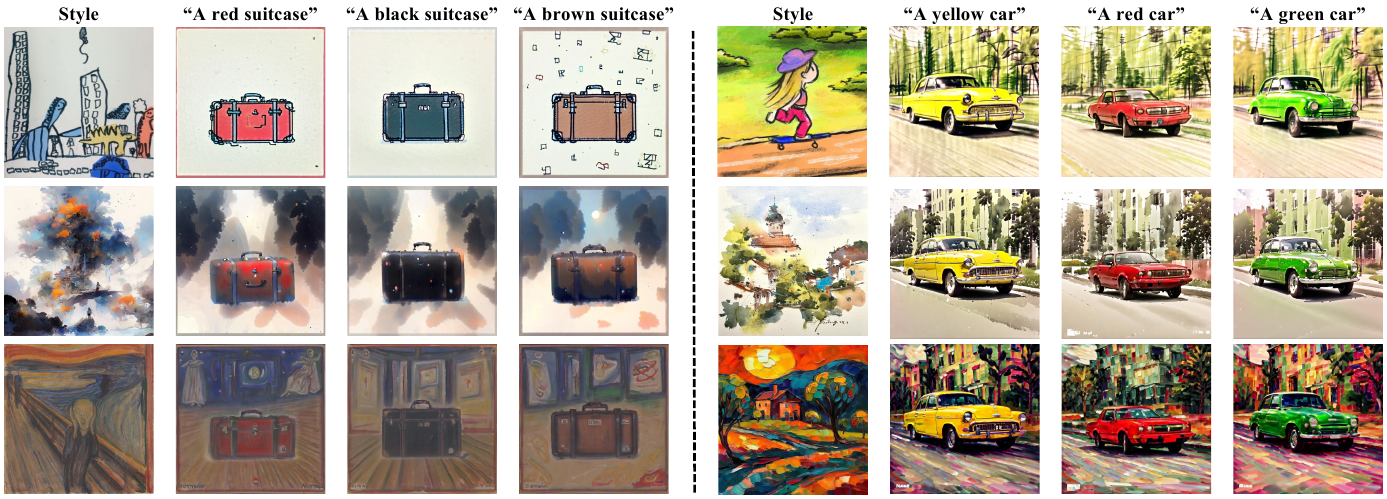}
    \vspace{-7mm}
    \captionof{figure}{Results of our text-driven style transfer model. Given a style reference image, our method effectively reduces style overfitting, generating images that faithfully align with the text prompt while maintaining consistent layout structure across varying styles.}
    \label{fig:teaser}
\end{center}
}]

\input{sec/0_abstract}
\input{sec/1_introduction}
\input{sec/2_related}
\input{sec/3_method}
\input{sec/4_results}
\input{sec/5_conclusion}
{
    \small

\input{main.bbl}
    \bibliographystyle{ieeenat_fullname}
}

\input{sec/6_suppl}

\end{document}

%% file: sec/0_abstract.tex
\begin{abstract}
Text-driven style transfer aims to merge the style of a reference image with content described by a text prompt.  Recent advancements in text-to-image models have improved the nuance of style transformations, yet significant challenges remain, particularly with overfitting to reference styles, limiting stylistic control, and misaligning with textual content.
In this paper, we propose three complementary strategies to address these issues. First, we introduce a cross-modal Adaptive Instance Normalization (AdaIN) mechanism for better integration of style and text features, enhancing alignment. Second, we develop a Style-based Classifier-Free Guidance (SCFG) approach that enables selective control over stylistic elements, reducing irrelevant influences. Finally, we incorporate a teacher model during early generation stages to stabilize spatial layouts and mitigate artifacts. 
Our extensive evaluations demonstrate significant improvements in style transfer quality and alignment with textual prompts.  Furthermore, our approach can be integrated into existing style transfer frameworks without fine-tuning.
\end{abstract}
\vspace{-7mm}
\blfootnote{\hspace{-2em}* denotes Corresponding author}

%% file: sec/1_introduction.tex
\section{Introduction}
\label{sec:1_introduction}

Text-driven style transfer is an important task in image synthesis, aiming to blend the style of a reference image with the content aligned to a text prompt.
Recent advancements in text-to-image generative models, such as Stable Diffusion~\cite{rombach2022high, podell2023sdxl, esser2024scaling}, have enabled nuanced style transformations pertaining to the reference image while preserving content fidelity. 
This technique holds significant practical value, particularly in domains such as digital painting, advertising, and game design. 

Nevertheless, modern style transfer techniques still fall short of expectations due to the inherent ambiguity in defining ``style.'' A style image encompasses various elements, including color palettes, textures, lighting, and brush strokes, all of which shape its overall aesthetic.
Existing models often replicate all these elements, which can inadvertently lead to overfitting, where the generated output overly mirrors the characteristics of the reference style image. 
This over-replication of details not only diminishes the aesthetic flexibility of the generated image but also restricts its adaptability to different stylistic or content-based requirements. Therefore, an ideal style transfer approach would thus allow for more selective stylistic adjustments, granting the user the flexibility to emphasize or omit specific stylistic components to achieve a balanced, intentional transformation.

\begin{figure}
    \centering
    \includegraphics[width=1.0\linewidth]{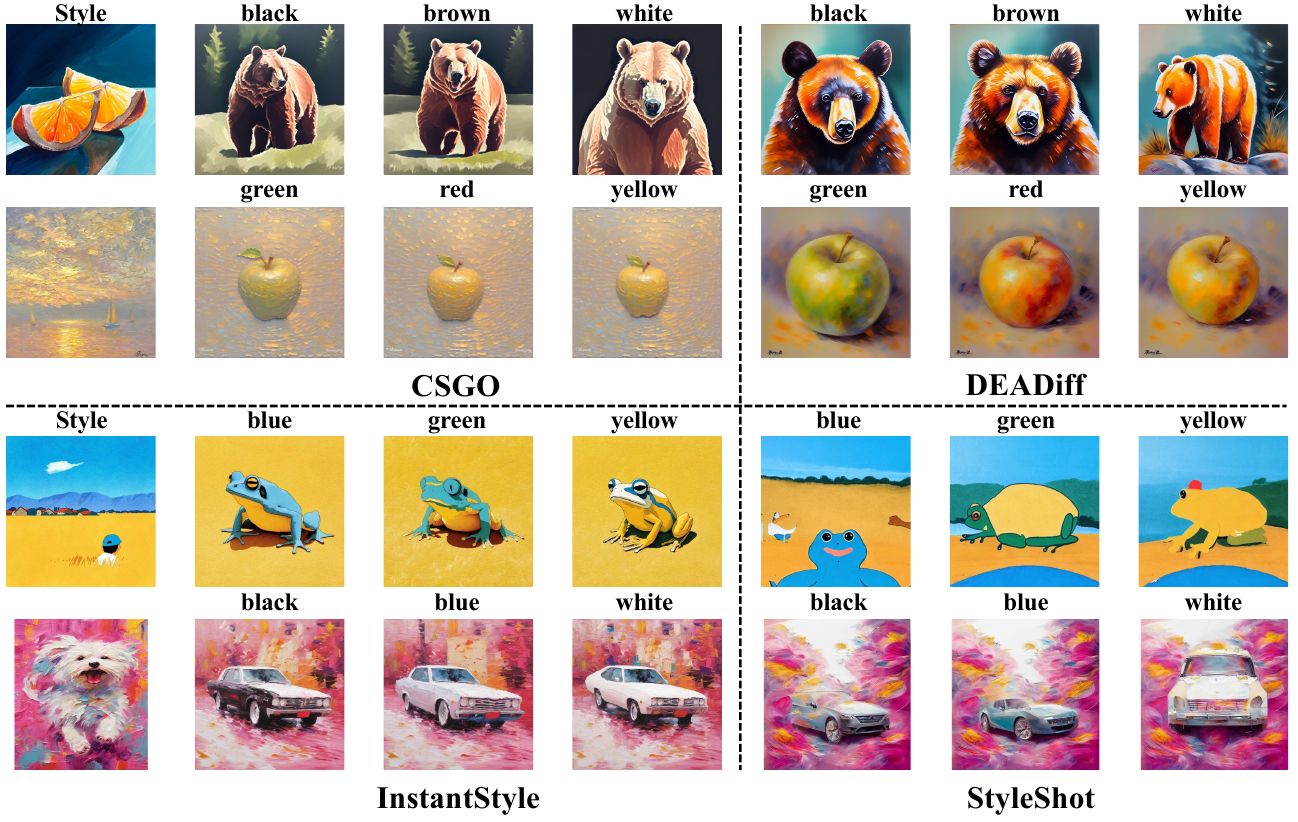}
    \vspace{-7mm}
    \caption{Illustration of overfitting issues in text-to-image generation, where the model tends to follow dominant colors or patterns from the style image rather than aligning precisely with the text prompt. Each prompt follows the format ``A $<$color$>$ $<$object$>$.'' From top to bottom, the objects are: bear, apple, frog, and car.}
    \label{fig:shortcome}
    \vspace{-3mm}
\end{figure}

Another challenge arising from overfitting is the difficulty in maintaining text alignment accuracy during text-to-image generation. As shown in \cref{fig:shortcome}, current models often prioritize dominant colors or patterns from the style image, even if they contradict instructions specified in the text prompt. This rigidity undermines the model’s ability to interpret and incorporate nuanced textual guidance, resulting in a decreased capacity for precision and customization in the generated output.
 
\begin{figure}
    \centering
    \includegraphics[width=1.0\linewidth]{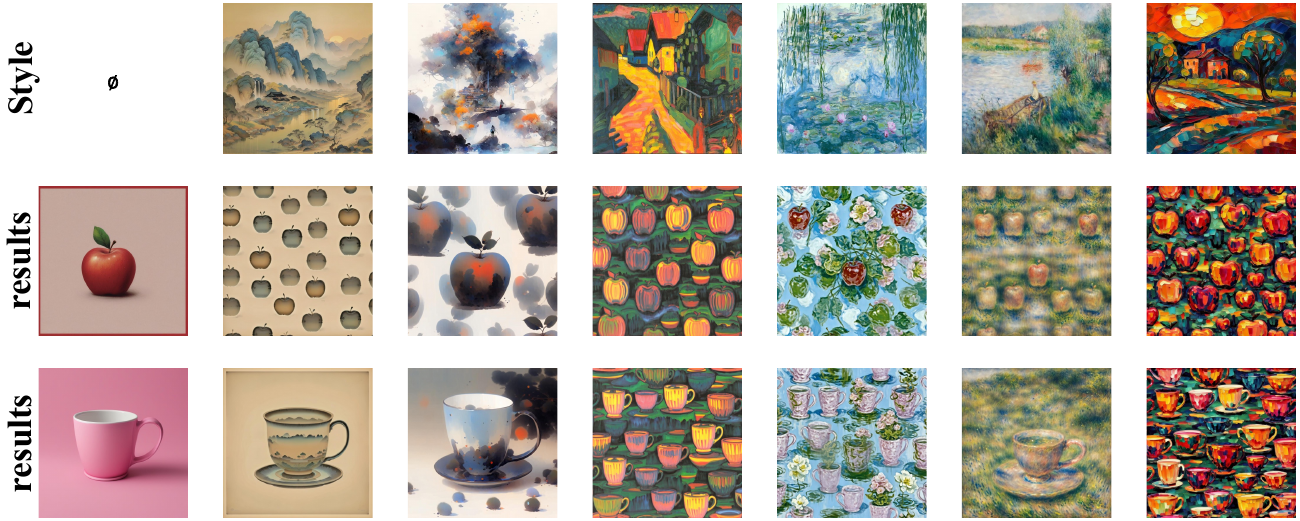}
    \vspace{-7mm}
    \caption{Illustration of the checkerboard artifact encountered in the CSGO~\cite{xing2024csgo} method during inference. The leftmost column shows the results generated by SDXL~\cite{podell2023sdxl}. The prompts, from top to bottom, are ``A red apple'' and ``A pink cup.'' All generated results use the same initial noise latent.}
    \label{fig:checkboard}
    \vspace{-5mm}
\end{figure}

Finally, style transfer can introduce undesirable artifacts, destabilizing the underlying text-to-image generation models.  One common artifact, as shown in \cref{fig:checkboard}, is the layout instability (\eg, checkerboard effect), wherein repetitive patterns inadvertently emerge throughout the generated image, irrespective of user instructions. This highlights the unique challenges introduced by the additional complexity of style transfer. 

In this paper, we propose three complementary strategies to address these challenges. 
First, to mitigate conflicts between the text prompt and the style reference during generation, we introduce a mechanism where the style image features are integrated into the text features using Adaptive Instance Normalization (AdaIN) before merging them with the image features. This adaptive integration creates a more cohesive guiding feature that subsequently guides the final image generation, aligning the stylistic features with the text-based instructions more harmoniously. 
Second, to disentangle and selectively control various elements within style images, we develop a style-based classifier-free guidance (SCFG) inspired by text-based classifier-free guidance in diffusion models. Specifically, we employ a layout-controlled generation model, such as ControlNet, to produce a comparable “negative” image that lacks the target style we aim to transfer. This negative image functions similarly to a ``null'' prompt in diffusion models, allowing the guidance to focus exclusively on the target style element and filter out extraneous stylistic features.
Finally, to enhance spatial stability, we incorporate a ``teacher model'' into the early stages of generation. The tutor model, based on the original text-to-image model, simultaneously performs denoising generation with the same textual prompt and shares its spatial attention maps with the style model at each time step. This method ensures stable and consistent spatial distribution, effectively mitigating issues like the checkerboard artifact. Additionally, this approach enables consistent spatial layouts across different style reference images for the same text prompt, facilitating more straightforward comparisons and evaluations of stylistic transformations.

In summary, our contributions are as follows:
\begin{itemize}
    \item AdaIN-based Integration: We develop cross-modal Adaptive Instance Normalization (AdaIN) to harmoniously integrate style and text features, improving alignment during generation.
    \item Style-based Classifier-Free Guidance (SCFG): We introduce a style-guided approach to focus on the target style and reduce unwanted stylistic features.
    \item Teacher Model for Layout Stability: We incorporate a teacher model to share spatial attention maps during early generation, ensuring stable layouts and mitigating artifacts like the checkerboard effect.
    \item Extensive evaluations conducted on a wide range of styles and prompts, as shown in \cref{fig:teaser}, demonstrate that our method significantly improves the alignment of the generated images. Furthermore, our approach is versatile and can be integrated into various existing style transfer frameworks while remaining fine-tuning-free.
\end{itemize}

%% file: sec/2_related.tex
\section{Related Work}
\label{sec/2_related}

\noindent\textbf{Text-to-image generation.}
Text-conditioned image generative models~\cite{rombach2022high, podell2023sdxl, saharia2022photorealistic, ramesh2022hierarchical, ding2021cogview, betker2023improving, dai2023emu} have demonstrated remarkable capabilities in generating high-quality images. Notably, models in the Stable Diffusion~\cite{rombach2022high, podell2023sdxl, esser2024scaling} series have achieved impressive advancements due to structural modifications and optimizations in their text encoder components. These improvements have led to significant enhancements in both the visual quality of generated images and the model's ability to align with complex textual prompts. The robust performance of text-to-image (T2I) diffusion models has also catalyzed the development of various related visual generation tasks, including text-based image editing~\cite{hertz2022prompt, liu2024towards, tumanyan2023plug, meng2021sdedit, mokady2023null, song2024doubly}, subject-driven image generation~\cite{jin2023image, ruiz2023dreambooth, gal2022image, nam2024dreammatcher, ding2024freecustom}, and other applications that leverage their strong generative and interpretative abilities.

\noindent\textbf{Stylized Image Generation.}
Style transfer applies the style of a reference image to a target content image and can be broadly categorized into image-driven and text-driven approaches. Image-driven methods, such as InST~\cite{zhang2023inversion}, StyleID~\cite{chung2024style}, and InjectFusion~\cite{jeong2024training}, focus on preserving content while injecting style, employing techniques like stochastic inversion, query preservation, and feature blending in the h-space~\cite{kwon2022diffusion}.
Text-driven methods aim to mitigate content leakage, where excessive style application distorts content features. B-LoRA~\cite{frenkel2025implicit} achieves style-content separation via LoRA weight optimization, while DEADiff~\cite{qi2024deadiff} enhances text-image consistency through joint cross-attention learning. StyleAlign~\cite{wu2021stylealign} and Visual Style Prompt~\cite{jeong2024visual} modify self-attention mechanisms by swapping either query-key (Q-K) or key-value (K-V) pairs, ensuring style alignment.
Adapter-based methods provide an efficient alternative by injecting style features into pre-trained diffusion models. IP-Adapter~\cite{ye2023ip} leverages a decoupled cross-attention mechanism for zero-shot style transfer and mitigates content leakage via weight tuning. InstantStyle~\cite{wang2024instantstyle} improves stylization by selectively injecting style features into Stable Diffusion’s UNet~\cite{rombach2022high, podell2023sdxl, ronneberger2015u}. StyleShot~\cite{gao2024styleshot} extracts multi-level style features for fine-grained control, while CSGO~\cite{xing2024csgo} enhances adapter-based stylization by training on a curated style dataset for effective style-content decoupling.

%% file: sec/3_method.tex
\section{Method}
\label{sec/3_method}
In this section, we detail the three complementary strategies we propose to address the challenges inherent in text-driven style transfer. Each strategy builds upon existing models \textbf{CSGO}~\cite{xing2024csgo} but introduces novel mechanisms to overcome the inherent limitations of current approaches in style transfer tasks.

\begin{figure*}
    \centering
    \includegraphics[width=1.0\linewidth]{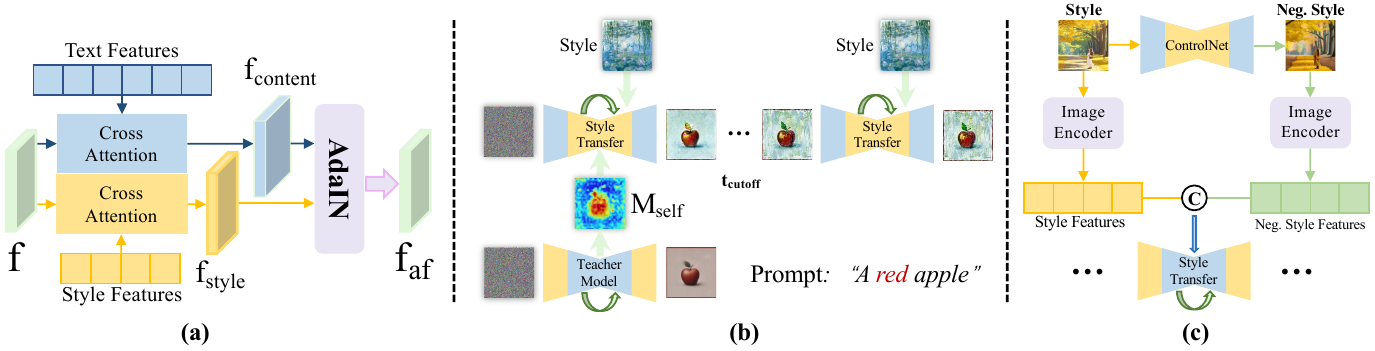}
    \vspace{-7mm}
    \caption{The illustration of our proposed Cross-Modal AdaIN, Teacher Model, Style-Based CFG.}
    \label{fig:allpipeline}
    \vspace{-7mm}
\end{figure*}

\subsection{Preliminaries}

Before delving into our specific methods, we begin by providing some background on the key components that underlie our approach.

\noindent\textbf{Latent Diffusion Models.}
Latent Diffusion Models (LDMs) ~\cite{rombach2022high} represent a powerful framework for efficient image generation. Operating in the latent space of a Variational Auto-Encoder (VAE)~\cite{kingma2013auto}, LDMs optimize computational efficiency while retaining high-quality image generation capabilities. Stable Diffusion (SD) \cite{rombach2022high, podell2023sdxl, esser2024scaling}, one of the most prominent models in this family, generates high-fidelity images by denoising noisy latent representations conditioned on text prompts. This denoising process can be formalized as:
\begin{equation}
    \label{eq:ldm}
    \mathcal{L}(\theta) = \mathbb{E}_{t, z, c, \epsilon \sim \mathcal{N}(0,1)}\left[\left\|\epsilon - \epsilon_{\theta}\left(z_{t}, t, c\right)\right\|_{2}^{2}\right],
\end{equation}
where $\epsilon$ is noise sampled from a standard Gaussian distribution, $\epsilon_{\theta}$ is the noise predicted by the U-Net architecture, $z_t$ is the noisy latent representation at timestep $t$, and $c$ is the conditioning text prompt. The goal of this loss function is to guide the model in iteratively reducing noise, which enables the model to generate high-quality images consistent with the input text.

\noindent\textbf{Attention Mechanisms.}
To enhance the generation process, Stable Diffusion incorporates several self-attention and cross-attention layers \cite{vaswani2017attention}. 
Recent studies show that self-attention captures dependencies within the latent representations, allowing the model to effectively gather context across different parts of the feature map. Cross-attention facilitates the integration of conditioning text embeddings with latent representations, ensuring that the generated output aligns with the input text prompt. This dual attention mechanism enhances the expressiveness and contextual relevance of the output, achieving a high level of quality and stylistic consistency.
The attention mechanism $A(\cdot)$, in its general form, can be defined as:
\begin{equation} 
    \label{eq} 
    \begin{split} 
   & Q = W_Q(f), \quad K = W_K(f'), \quad V = W_V(f'), \\
   &A(Q,K,V) = M V,\ \text{where}\  M = \text{softmax}(\frac{Q K^{T}}{\sqrt{d}}).
    \end{split} 
\end{equation}
$f$ and $d$ denote the input features and the dimensionality of head. For self-attention, $f' = f$, allowing the model to focus on intra-feature relationships. For cross-attention, $f'$ corresponds to the conditioning input, such as text embeddings, allowing the latent feature map $f$ to be modulated by the conditioning information. The attention map $M$ determines the focus of the model during generation.

\noindent\textbf{Style Transfer with an Adapter.}
Recent advancements in style transfer have explored the use of adapter-based methods to inject style-specific features into text-to-image models. These adapter networks are designed to modify pre-trained models to accommodate new styles inputs.
One notable approach is the IP-Adapter~\cite{ye2023ip}, which leverages a dual cross-attention mechanism to integrate both text and image conditions. The CSGO~\cite{xing2024csgo} shares the same architecture for text-driven style transfer, with the key distinction being that it is specifically trained on a style transfer dataset, enabling it to better capture and transfer stylistic features. In this design, one cross-attention layer extracts features from the text prompt, while the other extracts features from the image. The two sets of features are then combined through a weighted sum, allowing the model to generate images that adhere to both the stylistic attributes specified by the text and the visual characteristics of the image.
This combination process can be formalized as:
\begin{equation}
    f_\mathsf{ip} = A(Q, K_t, V_t) + \lambda A(Q, K_i, V_i),
    \label{eq:ipadapter}
\end{equation}
where $\lambda$ is a weight balancing the contributions from text-based and image-based attention. In this context, $Q$ is derived from the feature map $f$ of the latent representation, $K_t$ and $V_t$ are projections from the text embeddings $f_t$, and $K_i$ and $V_i$ are projections from the image feature map $f_i$.

While the IP-Adapter provides a straightforward and effective method for adapting models to perform style transfer, it has notable limitations. Specifically, the additive fusion of text and image conditions can lead to overfitting to the image style, particularly when there is conflicting information between the text and image conditions. In such cases, the model may overly prioritize the image style at the expense of the text prompt, resulting in outputs that are not aligned with the intended style or semantic content of the text. Furthermore, simply reducing the weight of the image condition can weaken the visual style, leading to suboptimal results. This introduces a challenge in determining an appropriate hyperparameter for balancing the contributions of both conditions.

In this paper, we introduce a new fusion strategy specifically tailored to style transfer tasks. Our method avoids the limitations of additive fusion by providing a more efficient and stable approach for combining text and image conditioning.

\subsection{Cross-Modal AdaIN of Text and Style Conditioning}
We propose a novel method for text-driven style transfer that better integrates both text and image conditioning. Our approach aims to achieve a balanced fusion of these two conditioning inputs, ensuring that they complement each other effectively.

In typical text-driven style transfer tasks, text conditioning primarily serves to define the content, dictating the semantic structure of the output, while image conditioning predominantly encodes the stylistic features, such as texture and color palette. However, directly combining these two conditioning inputs through weighted summation, as in existing methods~\cite{ye2023ip, wang2024instantstyle, xing2024csgo, gao2024styleshot}, forces them to assume similar roles in the fusion process. This can create conflicts, particularly when the text and image provide divergent information about the content and style. For example, a text prompt may describe a scene in one way, while the image may impose a stylistic choice that conflicts with this description, leading to suboptimal results.
To address this challenge, we revisit Adaptive Instance Normalization (AdaIN)~\cite{huang2017arbitrary}, a widely recognized technique in style transfer. AdaIN operates by normalizing the content input $x$ based on the statistical properties (mean and standard deviation) of the style input $y$, integrating style characteristics while preserving the essential structure of the content. The AdaIN process is defined as follows:
\begin{equation}
    \label{eq:adain}
    \textrm{AdaIN}(x, y) = \sigma(y) \left( \frac{x - \mu(x)}{\sigma(x)} \right) + \mu(y),
\end{equation}
where $\mu(x)$ and $\sigma(x)$ denote the mean and standard deviation of the content input $x$, and $\mu(y)$ and $\sigma(y)$ represent the mean and standard deviation of the style input $y$. By adjusting the content features to reflect the style statistics, AdaIN effectively fuses style into the content in a controlled manner, preserving the content’s alignment with the text description while ensuring stylistic consistency.

As shown in ~\cref{fig:allpipeline}(a), building on AdaIN, we develop a Cross-Modal AdaIN mechanism that integrates text and style conditioning in a way that respects their distinct roles. 
The Adapter-Based methods~\cite{ye2023ip, xing2024csgo} employed a weighted sum approach for feature fusion, ensuring that feature maps operated within the same embedding space. We discovered that AdaIN could directly replace the weighted sum strategy. This substitution enables effective feature fusion without the need for additional training, making it particularly advantageous when integrated with Adapter-Based methods~\cite{wang2024instantstyle, xing2024csgo}. Specifically, we first leverage the cross-attention layers within the U-Net architecture to query both the style and text features based on the U-Net feature map.
This step results in two separate grid feature maps—one for the style and one for the text—that share the same spatial resolution.
Next, we apply AdaIN between two feature maps, where the feature maps queried by text conditions are normalized by style feature maps.
Finally, the normalized text feature maps are fused into the raw U-Net features by simple addition. This fusion is constructed in a residual design, similar to the approach used in the raw cross-attention layers.
Mathematically, this adaptive fusion process can be expressed as follows:
\begin{equation} 
\label{eq:cross-modal-adain} 
    \begin{aligned} 
    f_{\text{style}} = A(Q, &K_{\text{style}}, V_{\text{style}}), 
    f_{\text{text}} = A(Q, K_{\text{text}}, V_{\text{text}}), \\ 
    \hat{f}_{\mathsf{af}} &= \gamma_{\text{style}} \cdot \left( \frac{f_{\text{text}} - \mu_{\text{text}}}{\sigma_{\text{text}}} \right) + \beta_{\text{style}}, \\
    \end{aligned} 
\end{equation}
where the $\gamma_{\text{style}}$ and $\beta_{\text{style}}$ come from the style image feature map, in the same way of run AdaIN.
By adaptively balancing the influence of text and style, our method effectively minimizes potential conflicts between the two inputs, eliminating the need for setting a tricky hyperparameter. 
\begin{figure}
    \centering
    \includegraphics[width=1.0\linewidth]{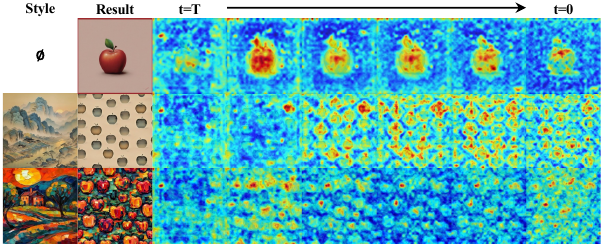}
    \vspace{-7mm}
    \caption{Visualization of the Cross-Attention Map for the word ``apple'' in the prompt ``A red apple'' during the generation process. When artifacts appear, the attention tends to scatter as well.}
    \label{fig:attnmap}
    \vspace{-5mm}
\end{figure}

\subsection{Layout Stablization with Teacher Model}

 In image generation, the layout is a crucial component of visual aesthetics. As shown in ~\cref{fig:checkboard}, we observe instances of artifacts during generation, such as checkerboard patterns. Upon analyzing the data presented in ~\cref{fig:attnmap}, we observed that these instabilities are correlated with a lack of aggregation in the core generative regions within the Cross-Attention mechanism. In the unstable generation examples, the layout instability reveal that the model fails to properly attend to regions associated with the word ``apple'' leading to visual distortions and compositional issues. This behavior diverges significantly from what is observed in the raw SDXL~\cite{podell2023sdxl} model at different timesteps.
 
In image generation, Self-Attention plays a crucial role in maintaining the layout and spatial structure of the original content~\cite{liu2024towards}. Self-Attention mechanism in Stable Diffusion~\cite{rombach2022high, podell2023sdxl} captures high-level spatial relationships, which effectively stabilize the foundational layout during generation. The preserved layout information, encapsulated within the Self-Attention AttnMaps, serves as a structural framework that guides the composition and distribution of elements across the image.

In the context of Text-Driven Style Transfer, maintaining a stable layout is crucial for ensuring that the generated image accurately reflects the structure and composition described by the textual prompt. Specifically, layout refers to the spatial arrangement of objects, elements, and the overall scene composition in the image. Without stable layout alignment, these elements may become misaligned, resulting in images where the content is distorted, or the intended focus, perspective, and balance outlined in the prompt are lost. To address this, we propose a method that stabilizes the layout by selectively replacing certain Self-Attention AttnMaps~\cite{vaswani2017attention, rombach2022high, podell2023sdxl} in the stylized image with those from the original diffusion model. This selective replacement helps preserve the spatial relationships and arrangement of key features in the image, ensuring that the core layout remains consistent throughout the denoising process. By doing so, we retain the structural coherence of the original image while still applying the desired stylistic transformation, leading to a more coherent and faithful alignment with the textual prompt. In~\cref{fig:allpipeline}(b), provides a visual overview of this process, showcasing the integration of layout stabilization within the generative framework.

Unlike conventional image editing approaches~\cite{tumanyan2023plug} that replace Self-Attention maps (AttnMaps) across all timesteps and decoder layers, our method selectively replaces AttnMaps only during the initial denoising timesteps but across all layers of the UNet. Applying full-timestep replacement, as conventional methods do, risks excessive loss of stylistic details, limiting the effectiveness of style transfer.

\subsection{Style-Based CFG}
A particularly challenging scenario in style transfer arises when the reference style image contains multiple stylistic elements, such as a combination of cartoon style and nighttime aesthetics. In such cases, the model faces style ambiguity, where various style features are present, but the focus is intended to be on just one specific element. Current methods struggle to effectively disentangle these different styles and selectively emphasize the desired one.
To address this challenge, a flexible method is required that can selectively emphasize the desired style elements while filtering out irrelevant or conflicting features.
Inspired by the concept of classifier-free guidance (CFG) \cite{ho2022classifier}, commonly used in diffusion models for text-guided image generation, we propose a Style-Based CFG design to provide controlled adjustments in the style transfer process. 

\begin{figure*}
    \centering
    \includegraphics[width=1.0\linewidth]{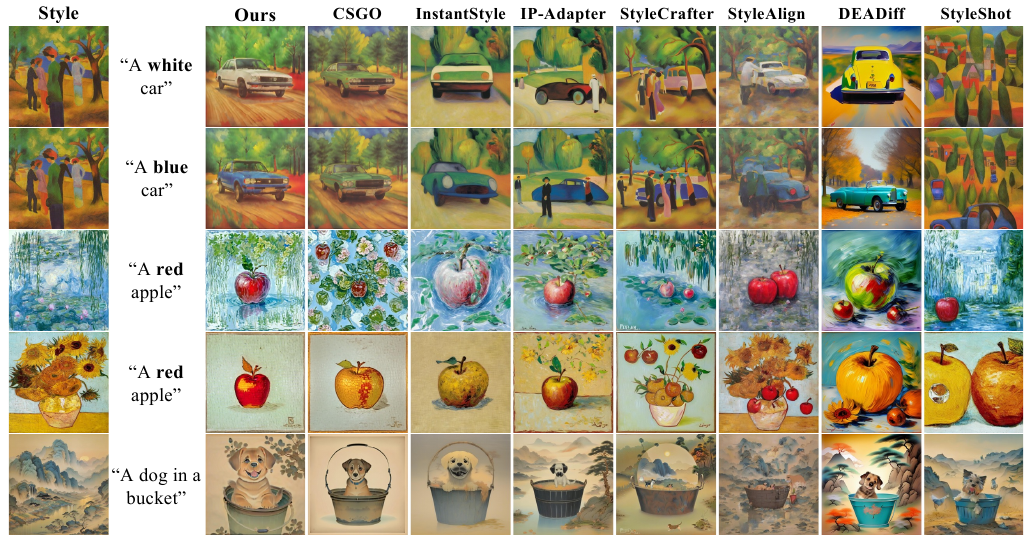}
    \vspace{-7mm}
    \caption{Qualitative comparison with state-of-the-art methods. Our approach effectively preserves image style while accurately adhering to text prompts for generation.}
    \label{fig:quality}
    \vspace{-3mm}
\end{figure*}

\noindent\textbf{Classifier-Free Guidance Mechanism.}
In standard CFG, the model generates outputs conditioned on a given text prompt, as well as outputs generated without any conditioning (i.e., the unconditional model output). The final output is a weighted combination of these two, where the conditional output steers the generation process in alignment with the prompt, while the unconditional output acts as a form of negative conditioning, helping to prevent the model from producing undesirable features.
The unconditional output in the Style-Based CFG can be replaced with conditioning on a negative prompt, such as "blur" or "artifact" to actively discourage the generation of undesirable features.
Therefore, the CFG~\cite{ho2022classifier} mechanism can be formalized as:
\begin{equation}
    \label{cfg}
    {\small
    \hat{\epsilon}_\theta(z_t, t,y) = (1 + w) \cdot \epsilon_\theta(z_t,t, y_\text{cond}) - w \cdot \epsilon_\theta(z_t,t,y_\text{neg}),}
\end{equation}
where $y_\text{cond}$ is the positive condition and $y_\text{neg}$ is the negative condition, and $w$ is a weight controlling the balance between these outputs. 

\noindent\textbf{Style-Based Classifier-Free Guidance.}
In the context of style transfer, we extend this CFG mechanism to address the challenge of style ambiguity in images. Specifically, we introduce the concept of a negative style image that retains the overall content of the reference image but excludes the target style element. This negative image serves as a counterpart to the original style image and helps the model focus on transferring only the desired style component.
To generate the negative style image, we use a layout-controlled generation model, such as ControlNet, which allows us to create an image $z_t^{\text{neg}}$ that preserves the structural features of the reference image but omits the target style. This negative image functions similarly to a negative prompt in text-guided CFG, effectively guiding the model to emphasize the desired style while suppressing undesired style elements.
As show in~\cref{fig:allpipeline}(c), the SCFG mechanism operates as follows: 1) {Generate a Negative Image}: Using ControlNet, generate a negative sample image $z_t^{\text{neg}}$ that retains the structural elements of the image while omitting the target style.
2) {Apply SCFG to Guide Generation}: Formulate the Style-Based CFG by modulating the balance between the target style image $z_t^{\text{style}}$ and the negative style image $z_t^{\text{neg}}$. This balance is governed by a weight factor $w$, which determines the contribution of each image during generation.
The noise prediction is modified as follows:
\begin{equation}
    \label{styleCFG}
    \begin{split}
    \hat{\epsilon}_\theta(z_t, t, y) &= (1 + w) \cdot \epsilon_\theta(z_t, y^{text}_\text{cond}, y^{style}_\text{cond}) \\ 
    & - w \cdot \epsilon_\theta(z_t, y^{text}_{neg}, y_{neg}^{style}),
    \end{split}
\end{equation}

By integrating SCFG, our method refines the generation process by isolating specific style components, filtering out extraneous ones, and thereby focusing style transfer on the desired features. This approach reduces the risk of overfitting to irrelevant style components, allowing the model to perform effective style transfer in complex scenarios with multiple stylistic elements.

%% file: sec/4_results.tex
\section{Evaluation and Experiments}
\label{sec/4_results}

\begin{table*}[htp]
\centering
\aboverulesep=0pt
\belowrulesep=0pt
\scriptsize
\begin{tabular}{l|ccccc|cc|c}
    \toprule
    \multirow{2}{*}{\textbf{Metric}} & \multicolumn{5}{c|}{SDXL-based Methods} & \multicolumn{2}{c|}{SD15-based Methods} & \multirow{2}{*}{\textbf{Ours}} \\
    
    \cline{2-8}& IP-Adapter~\cite{ye2023ip} & InstantStyle~\cite{wang2024instantstyle} & CSGO~\cite{xing2024csgo} & StyleAlign~\cite{wu2021stylealign} & StyleCrafter~\cite{liu2023stylecrafter} & StyleShot~\cite{gao2024styleshot} & DEADiff~\cite{qi2024deadiff} \\
    \hline
    \midrule
    Text Alignment $\uparrow$ & 0.221 & 0.229 & 0.216 & 0.180 & 0.189 & 0.202 & 0.229 & \textbf{0.235} \\
    infer Time $(s)$ & 6 & 6 & 9 & 48 & 4 & 3 & 2 & 17 \\
    User-study Text $\%$ & 7.48 & 6.46 & 7.99 & 5.78 & 3.06 & 2.55 & 1.87 & \textbf{62.92} \\
    User-study Style $\%$ & 6.63 & 8.67 & 6.97 & 7.82 & 8.67 & 5.10 & 5.27 & \textbf{50.85} \\
    \bottomrule
\end{tabular}
\caption{Quantitative comparison with state-of-the-art methods. Our approach achieves the best performance on the text alignment metric and outperforms others in the user study evaluation.}
\label{quantity:styleadapterPrompt}
\vspace{-5mm}
\end{table*}

\begin{table}[htp]
\centering
\aboverulesep=0pt
\belowrulesep=0pt
\scriptsize
\begin{tabular}{cc|l} 
    \toprule
    Cross-Modal AdaIN & Teacher Model  & Text Alignment $\uparrow$ \\
    \midrule
     & & 0.216 \\
     & \checkmark & 0.223 \textcolor{green}{(+3.2\%)} \\
    \checkmark & & 0.228 \textcolor{green}{(+5.5\%)} \\
    \checkmark & \checkmark & 0.235 \textcolor{green}{(+8.7\%)} \\
    \bottomrule
\end{tabular}
\caption{Ablation study evaluating the impact of our proposed methods. Both designs significantly enhance text alignment accuracy.}
\label{ablation:styleadapterPrompt}
\vspace{-5mm}
\end{table}

\textbf{Implementation details.}
We have implemented our method on top of the latest Adapter-Based Style Transfer approach, named CSGO~\cite{xing2024csgo}. To ensure fairness in comparison and mitigate the influence of random initialization, we fixed the initial noise for all methods, as the initial noise can significantly impact the final results~\cite{hertz2022prompt}. 

\noindent\textbf{Evaluation.}
To enable a comparison with existing methods, we evaluated both our approach and several previous ones, including CSGO~\cite{xing2024csgo}, InstantStyle~\cite{wang2024instantstyle}, IP-Adapter~\cite{ye2023ip} with weight tuning, StyleCrafter~\cite{liu2023stylecrafter}, StyleAlign~\cite{wu2021stylealign}, DEADiff~\cite{qi2024deadiff}, StyleShot~\cite{gao2024styleshot}. To evaluate the performance of these methods in terms of prompt adherence after text-driven style transfer, we constructed a benchmark consisting of 52 prompts and 20 style reference images. These prompts were selected from the settings used in StyleAdapter~\cite{wu2021stylealign},while the style images are taken from StyleShot~\cite{gao2024styleshot}. Additionally, using ChatGPT, we generated 30 prompts in the form of ``A $<$color$>$ $<$object$>$'' which better highlights the issue of style overfitting.

More results of our method can be found in~\cref{sec:more_result}, and its integration with other methods is detailed in~\cref{sec:integration}.

\begin{figure}
    \centering
    \includegraphics[width=1.0\linewidth]{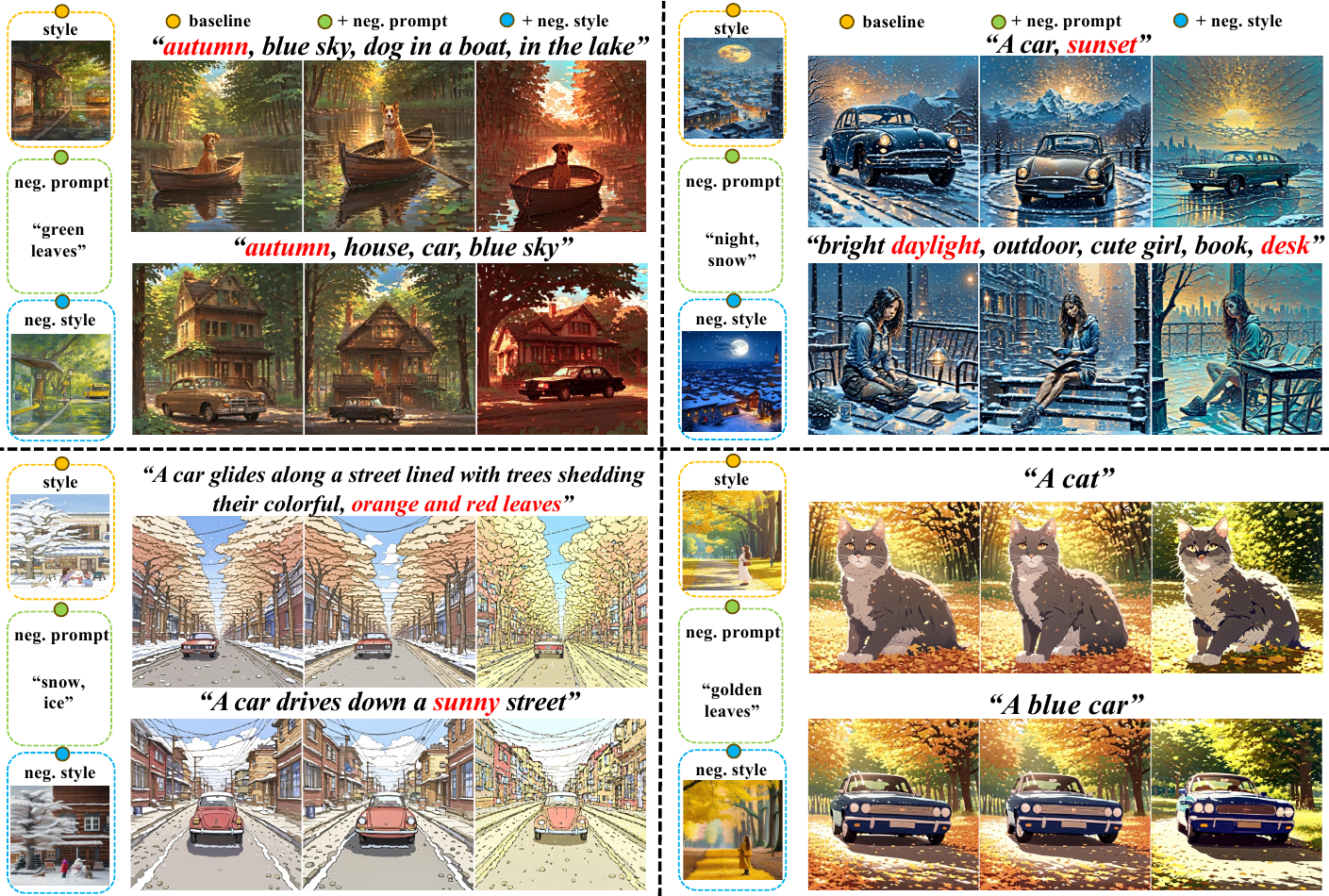}
    \vspace{-7mm}
    \caption{Impact of Style-Based CFG. The proposed Style-Based CFG successfully eliminates unintended style elements such as ``green leaves'', ``night'', ``snow'' and ``golden leaves'', while text-based CFG  fails to address these spurious attributes.}
    \label{fig:styleCFG}
    \vspace{-5mm}
\end{figure}

\subsection{Comparison with State-of-the-Arts}
\textbf{Qualitative Comparisons.}
\cref{fig:quality} presents qualitative comparisons with state-of-the-art methods. Existing approaches struggle with prompt alignment due to style overfitting. While CSGO~\cite{xing2024csgo}, InstantStyle~\cite{wang2024instantstyle}, and DEADiff~\cite{qi2024deadiff} mitigate content leakage, they fail to fully capture prompt-specified details. Conversely, IP-Adapter~\cite{ye2023ip}, StyleShot~\cite{gao2024styleshot} and StyleCrafter~\cite{liu2023stylecrafter} exhibit content leakage, leading to images misaligned with the input prompts. In contrast, our method ensures better prompt alignment while maintaining layout stability.
More detailed results and explanations could be found in~\cref{sec:Add_Compare}.

\noindent\textbf{Quantitative Comparisons.}
To verify the alignment between the generated image and its specified object, we compute the CLIP cosine similarity~\cite{radford2021learning} between the image and the corresponding text description.
As shown in \cref{quantity:styleadapterPrompt}, Our method outperforms the others, achieving the highest text alignment capability.

\noindent\textbf{User Study.}
We also conducted a user study to assess users' evaluations of text alignment and style similarity, with the results presented in ~\cref{quantity:styleadapterPrompt}. For each method's generated images, 49 users participated in an anonymous vote, selecting the example they felt was the most aligned with the text description and the closest in style to the reference image. The normalized votes (vote rate) serve as the scores for text alignment and style similarity.

\subsection{Style-Based CFG}
We conducted experiments on Style-Based CFG (SCFG), with the results shown in ~\cref{fig:styleCFG}. In the baseline images, unintended style elements such as snow and golden leaves are present, which do not align with the intended style. Adding a negative text prompt allowed for some control over these elements; however, unintended style elements like snow were not effectively mitigated. By applying a negative style image, we achieved more effective control, successfully removing these unwanted elements. This demonstrates the effectiveness of SCFG in precisely managing unintended style elements in generated images.

\subsection{Ablation Study}
\textbf{Cross-Modal AdaIN.} To demonstrate the effectiveness of each component in our method, we conducted an ablation study focusing on a quantitative analysis of text alignment. The same dataset used in the quantitative evaluation was employed for this test.
As shown in \cref{ablation:styleadapterPrompt}, our method consistently improves text alignment accuracy over the baseline. Incorporating the Teacher Model alone provides an initial enhancement, while cross-modal AdaIN yields a more significant improvement. Finally, combining both the cross-modal AdaIN and the Teacher Model results in the highest level of improvement, indicating complementary effects that enhance alignment performance.

\begin{figure}
    \centering
    \includegraphics[width=1.0\linewidth]{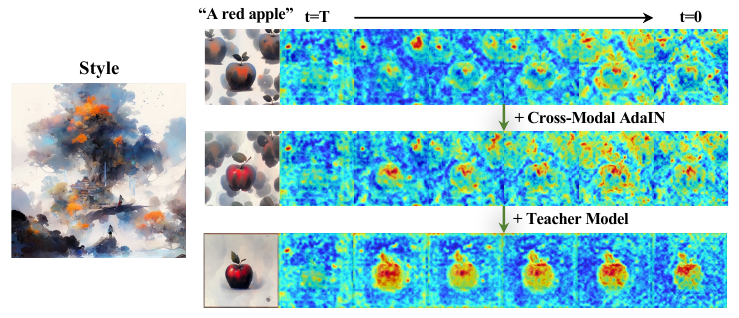}
    \vspace{-7mm}
    \caption{Visualization of cross-attention maps for the word ``apple'' in the prompt ``A red apple'' across different models. The proposed teacher model effectively rectifies attention maps, leading to improved image generation quality.}
    \label{fig:abl_teachermodel}
    \vspace{-3mm}
\end{figure}

\begin{figure}
    \centering
    \includegraphics[width=1.0\linewidth]{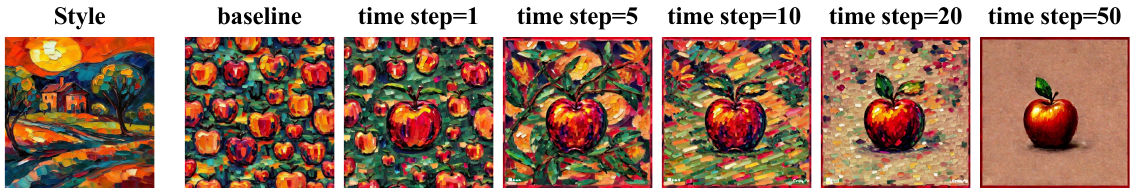}
    \vspace{-7mm}
    \caption{Impact of Teacher Model on Style Image Generation. The term ``timestep'' refers to the number of denoising steps during which the teacher model is involved.}
    \label{fig:abl_teachertime}
    \vspace{-5mm}
\end{figure}

\noindent\textbf{Teacher Model.} In \cref{fig:abl_teachermodel}, shows a visualization of cross-attention maps for the word ``apple'' in the prompt ``A red apple'' across different model configurations. The first row represents the baseline model, where the attention map fails to effectively focus on the core concept of ``apple'' from the text prompt. In the second row, the addition of cross-modal AdaIN provides some improvement, but the attention is still diffuse and lacks precise focus. In the third row, with the Teacher Model added, the attention map becomes more concentrated on the target concept, demonstrating a clear improvement in alignment with the prompt. This comparison indicates that it is the Teacher Model that enables the model to better capture and emphasize key elements from the text prompt, resulting in improved attention quality and image generation accuracy.
In \cref{fig:abl_teachertime}, this experiment demonstrates the importance of selecting an appropriate denoising timestep to stop the involvement of the Teacher Model. If the Teacher Model’s influence is removed too early, as seen at lower timesteps, issues with layout stability persist, resulting in compositions that lack coherence and include multiple instances of the target object (``apple''). However, if the Teacher Model is involved throughout the full denoising process (\eg, at timestep 50), there is a noticeable loss of style in the generated image. This suggests that while the Teacher Model is crucial for achieving layout stability, an extended involvement can dilute the style elements, making it essential to find a balanced point to discontinue its influence for optimal results.
More detailed results and explanations can be found in~\cref{sec:ablation}.

\begin{figure}[H]
  \centering
  \vspace{-2mm}
  \includegraphics[width=1.0\linewidth]{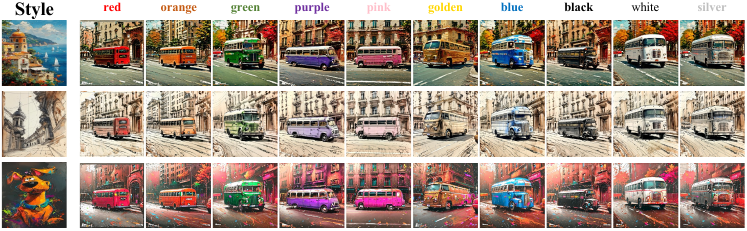}
  \vspace{-7mm}
  \caption{More results of our text-driven style transfer model. Illustration of the prompt format used: ``A [color] bus".}
  \label{fig:ours_bus_small}
  \vspace{-6mm}
\end{figure}

\begin{figure}[H]
  \centering
  \includegraphics[width=1.0\linewidth]{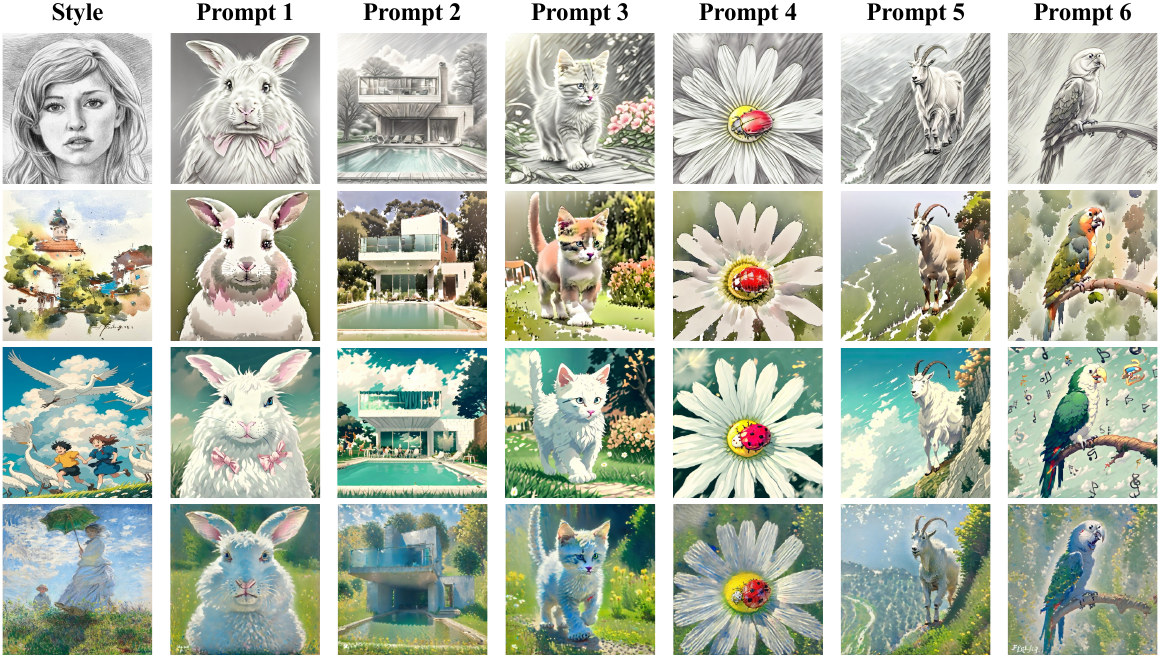}
  \vspace{-7mm}
  \caption{The prompts, from left to right, are: ``A fluffy white rabbit with pink ears and nose", ``A modern house with a pool.", ``A lovely kitten walking in a garden.", ``A daisy with a ladybug on it", ``A mountain goat on a cliff", and ``A parrot singing a song".}
  \label{fig:complex_ours}
  \vspace{-5mm}
\end{figure}

\noindent\textbf{More Results.} ~\cref{fig:ours_bus_small} demonstrates that our method effectively mitigates style overfitting by consistently adapting the target color while preserving structural details. Similarly, ~\cref{fig:complex_ours} shows that our model can handle complex prompts without losing style fidelity or scene composition. More detailed results can be found in~\cref{sec:more_result}.

%% file: sec/5_conclusion.tex
\section{Conclusion}
\label{sec/5_conclusion}

In conclusion, existing text-driven style transfer faces key issues such as style overfitting and layout instability, which limit the adaptability and coherence of generated images. To address these challenges, we proposed three methods: cross-modal AdaIN for harmonizing style and text features, style-based classifier-free guidance (SCFG) for selective control of stylistic elements, and a Teacher Model to enhance layout stability. Our results confirm that our approach effectively mitigates these issues, improving alignment, stability, and control in style transfer, making it a versatile and robust solution for text-to-image synthesis tasks.

\noindent\textbf{Limitations and Future Work.}
While our method improves artifact removal and layout stability, the Teacher Model slightly increases inference time. Additionally, generating negative-style image requires expertise and manual effort. Future work could focus on improving efficiency and further exploring strategies to mitigate style overfitting, enabling more adaptive and generalizable style transfer across diverse prompts and visual domains.

%% file: sec/6_suppl.tex
\label{sec/6_suppl}

\clearpage
\setcounter{page}{1}

\appendix
\newcommand{\AppendixPrefix}{A}

\noindent This \textbf{Appendix} provide additional details regarding the experimental setup described in the main paper and offer an extended analysis of the contributions of individual components. The content is organized as follows:
\begin{itemize}
\item \textbf{Details of Experiments.} This section provides additional information about the experiments discussed in the main paper, including specifics on the quantitative evaluations and the user study setup.
\item \textbf{Ablation Study.} Qualitative comparisons from the ablation experiments are presented, analyzing the impact of the Teacher Model, particularly in terms of timestep selection and the choice of Attention Map.
\item \textbf{Additional Qualitative Comparisons.} This section presents extensive qualitative comparisons, demonstrating that cross-modal AdaIN effectively prevents style overfitting, while the Teacher Model ensures layout stability and mitigates the occurrence of artifacts.
\item \textbf{Integration with Other Methods.} This section explores how our approach can be integrated with existing methods, such as InstantStyle~\cite{wang2024instantstyle} and StyleCrafter~\cite{liu2023stylecrafter}, showcasing its ability to enhance their performance and adaptability.
\end{itemize}

\section{Implementation Details}
We set the random seed to 42 for reproducibility, used 50 inference steps, and applied a uniform guidance scale of 5 across all methods. In the qualitative and quantitative comparison experiments, for the implementation involving the Teacher Model, its participation was specifically limited to the first 20 steps. All experiments were conducted on a single NVIDIA GTX-4090 GPU.

Adapter-based methods~\cite{ye2023ip, wang2024instantstyle, xing2024csgo, gao2024styleshot} are particularly suitable for style transfer. Their fine-tuning-free nature, combined with high-quality style transfer performance, has made them widely adopted. CSGO~\cite{xing2024csgo} employs a widely used adapter-based model structure and is the first method trained on a meticulously curated dataset specifically designed for style transfer. This effectively decouples the content and style in style images, enhancing the grasp of style details such as brushstrokes and textures. Therefore, in the experimental section, we selected it as the baseline and implemented specific modifications based on it. The implementation details are as follows:

We only retained the modules in CSGO~\cite{xing2024csgo} related to text-driven style transfer, removing irrelevant components, \eg, ControlNet~\cite{zhang2023adding}. This optimization reduces potential interference while lowering experimental costs, including memory usage during inference. At the same time, both the Teacher Model and cross-modal AdaIN are optional and can be used based on specific needs. For the quantitative experiments in the main paper, we incorporated both the Teacher Model and the cross-modal AdaIN module to achieve optimal text alignment. In the qualitative and quantitative comparison experiments, the Teacher Model participated for the first 20 time steps, with the total number of inference steps set to 50.

\begin{algorithm}[!ht]
\small
    \renewcommand{\algorithmicrequire}{\textbf{Input:}}
    \renewcommand{\algorithmicensure}{\textbf{Output:}}
    \caption{SDXL-Guided Self-Attention Replacement}
    \label{SGSAR}
    \begin{algorithmic}[1] 
        \REQUIRE  $P_\text{dst}$: a target prompt;
                  $I_\text{ref}$: style reference image;
                  $S$: random seed;
                  \textrm{DM}: raw Stable Diffusion Model;
                  \textrm{ST}: style transfer Method Model;
                  $t_\text{cutoff}$: stop replacement time step;
        \ENSURE   $I_\text{style}$: text-driven stylized image;
        
        \STATE $z_{T} \sim \mathcal{N}(0,1)$, a unit Gaussian random value sampled with random seed $S$;
        \STATE $z^{*}_{T} \leftarrow z_{T}$;
        
        \FOR {$t = T, T-1, \ldots, 1$}
            \IF{$t > t_\text{cutoff}$}
                \STATE $z_{t-1}, M_{self} \leftarrow \textrm{DM}(z_{t}, P_{dst}, t)$; 
                \STATE $z^{*}_{t-1} \leftarrow \textrm{ST}(z^{*}_{t}, I_\text{ref}, P_\text{dst}, t)\{M^{*}_\text{self} \leftarrow M_\text{self}\}$; 
            \ELSE
                \STATE $z^{*}_{t-1} \leftarrow \textrm{ST}(z^{*}_{t}, I_\text{ref}, P_\text{dst}, t)$
            \ENDIF
        \ENDFOR
        
        \STATE \textbf{Return} $I_\text{res} \leftarrow \text{Decoder}(z_{0})$;
    \end{algorithmic}
\end{algorithm}

\section{Evaluation Settings and User Study}
\begin{figure}
    \centering
    \includegraphics[width=1.0\linewidth]{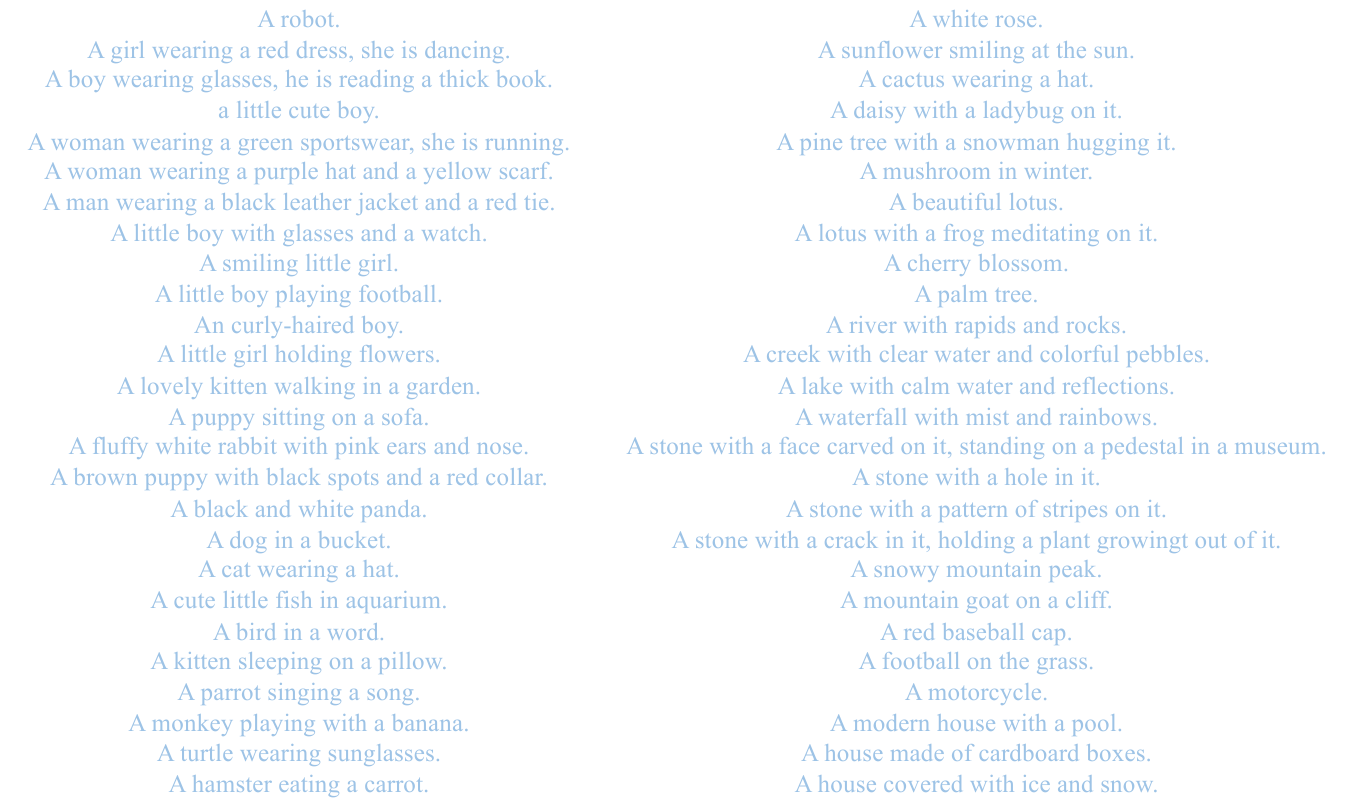}
    \vspace{-7mm}
    \caption{Details of the Test Set. The prompts used in the quantitative experiments were derived from StyleAdapter~\cite{wang2023styleadapter}.}
    \label{fig:supp_prompts}
    \vspace{-5mm}
\end{figure}

\begin{figure}
    \centering
    \includegraphics[width=1.0\linewidth]{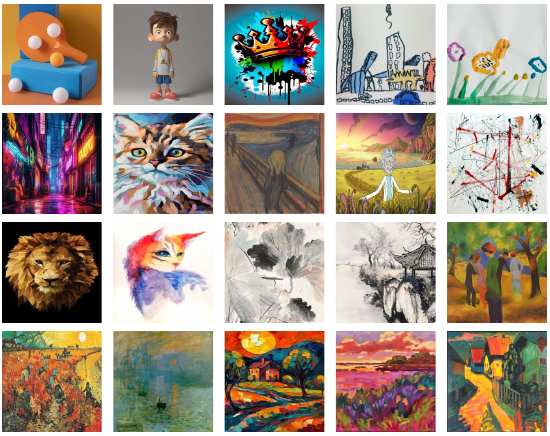}
    \vspace{-7mm}
    \caption{Details of the Test Set. The style images used in the quantitative experiments were randomly sampled from the test set of StyleShot~\cite{gao2024styleshot}.}
    \label{fig:supp_style_ref}
    \vspace{-5mm}
\end{figure}

In the quantitative experiments presented in the main paper, the evaluation was conducted using prompts derived from StyleAdapter~\cite{wang2023styleadapter}, with specific examples provided in~\cref{fig:supp_prompts}. The style images were randomly sampled from the test set of StyleShot~\cite{gao2024styleshot}, with representative examples shown in~\cref{fig:supp_style_ref}. Ultimately, each method generated 1,000 images for the quantitative experiments.

\begin{figure*}
    \centering
    \includegraphics[width=1.0\linewidth]{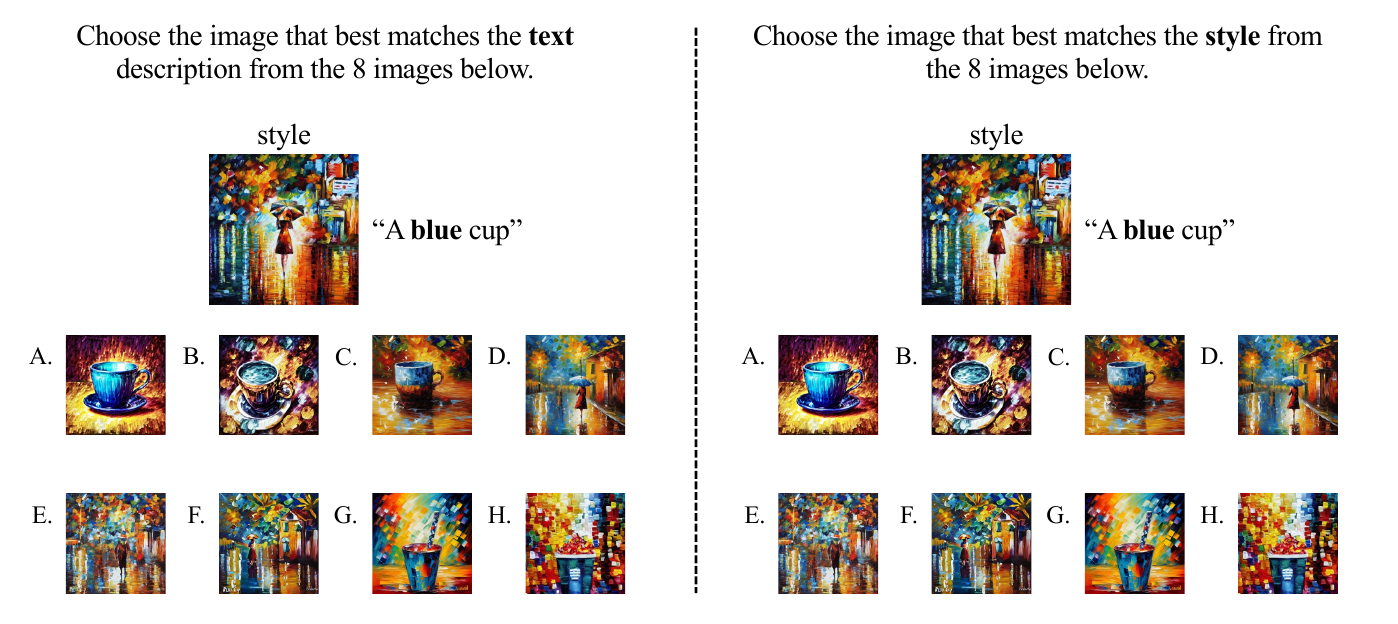}
    \vspace{-7mm}
    \caption{The questionnaire format for the user study. Each option represents the generation result of a method under a given style and prompt.}
    \label{fig:supp_user_study}
    \vspace{-5mm}
\end{figure*}

Beyond quantitative evaluations, we conducted a user study to gain subjective insights into the performance of different methods. The study involved 12 pairs of reference images and prompts. For each pair, participants were asked to assess and select the method they found superior based on two criteria: text alignment and style similarity. To ensure a fair assessment, participants were provided with a brief explanation of the task and evaluation criteria beforehand. We collected responses from 49 participants with diverse backgrounds, including individuals with relevant expertise in text-to-image tasks. The specific design of the questionnaire, including example pairs and evaluation guidelines, is shown in ~\cref{fig:supp_user_study}.

\begin{figure*}
    \centering
    \includegraphics[width=1.0\linewidth]{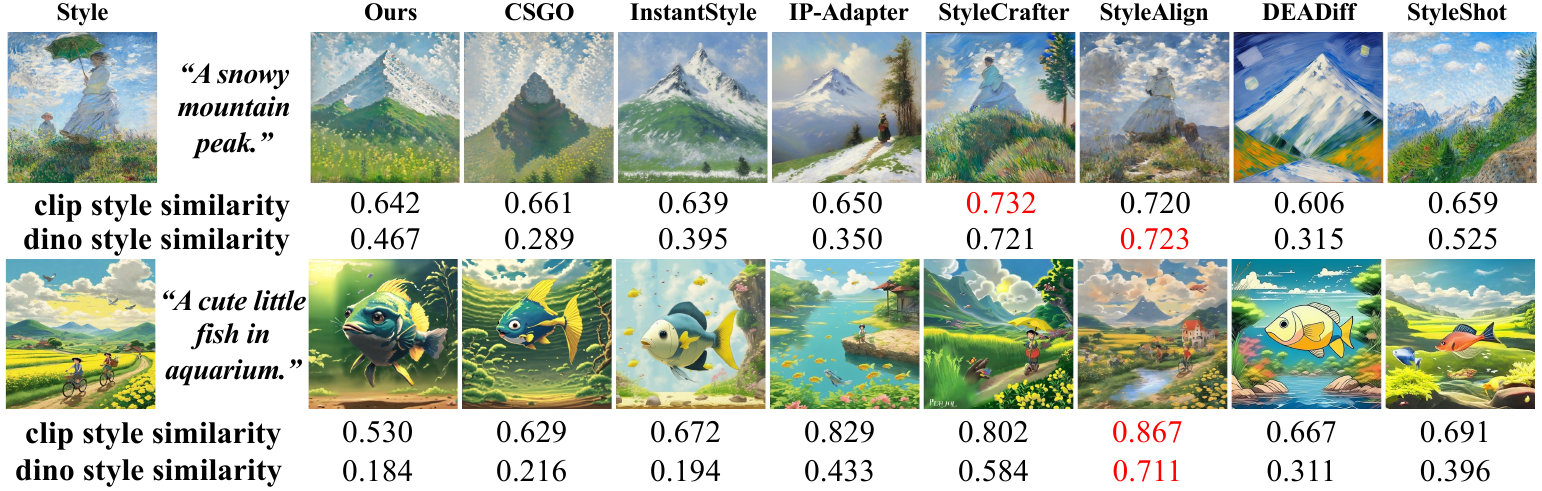}
    \vspace{-7mm}
    \caption{We observed that existing metrics generally fail to capture adherence to style. They tend to favor higher semantic similarity to the style image rather than better style transfer, a known issue often referred to as content leakage. A higher semantic similarity score does not indicate better style preservation and can, in fact, weaken the style in the generated results.}
    \label{fig:supp_style_metrics}
\end{figure*}

\begin{figure*}
    \centering
    \includegraphics[width=1.0\linewidth]{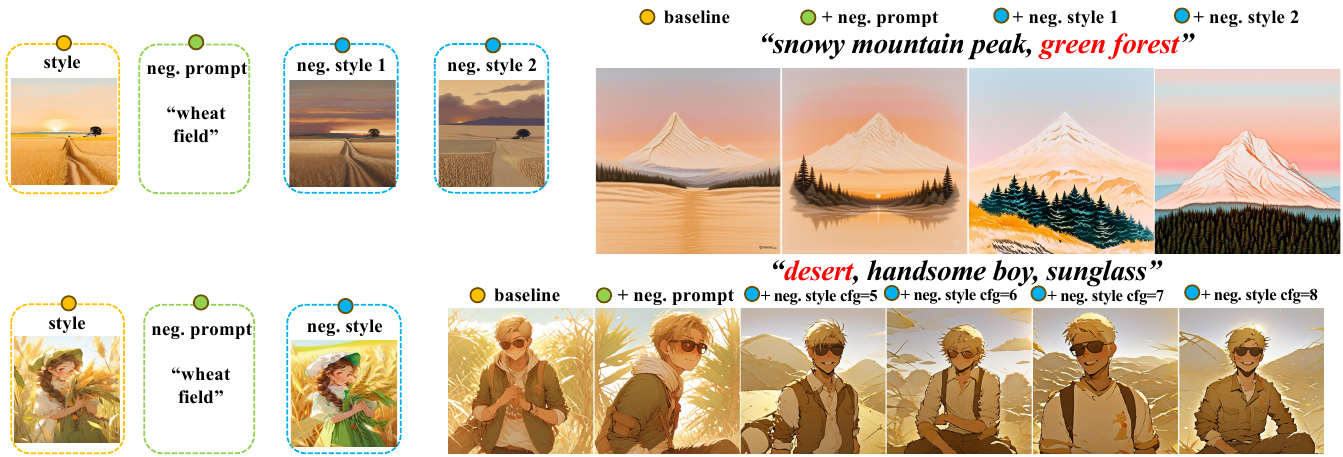}
    \vspace{-5mm}
    \caption{More results of Style-Based CFG.}
    \label{fig:supp_scfg_more}
    \vspace{-5mm}
\end{figure*}

\section{Additional Ablation Study}
\label{sec:ablation}
\begin{figure*}
    \centering
    \includegraphics[width=1.0\linewidth]{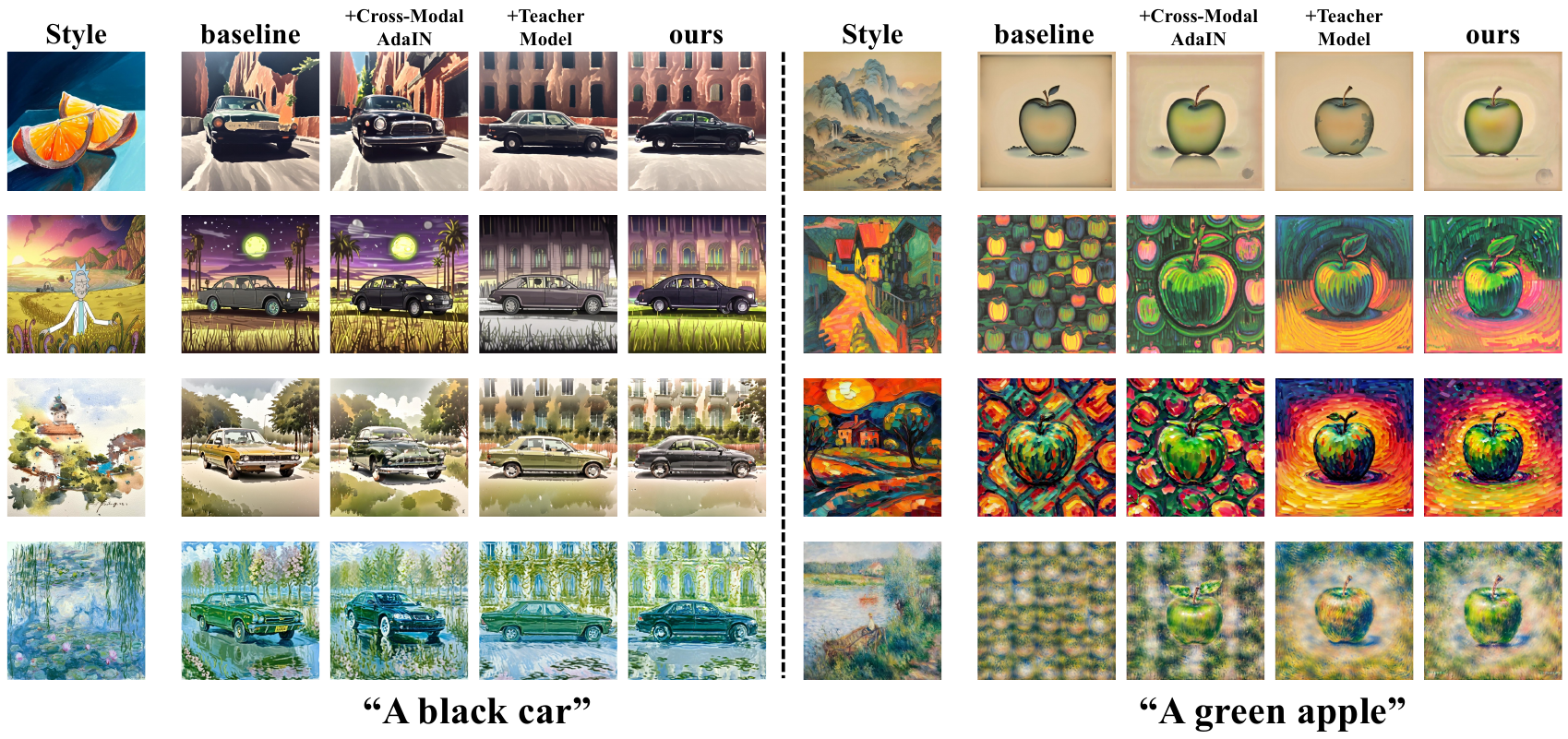}
    \vspace{-7mm}
    \caption{Qualitative results of the ablation study. cross-modal AdaIN enhances text alignment while preserving style similarity, addressing style overfitting issues. Incorporating the Teacher Model improves layout stability and resolves artifacts, ensuring consistent layout arrangements across different styles, as demonstrated in the ``A green apple'' example.}
    \label{fig:supp_ablation_1}
    \vspace{-5mm}
\end{figure*}

\textbf{Qualitative Results of the Ablation Study.} While the main paper presents a quantitative analysis, a qualitative comparison provides a more intuitive understanding of the contributions of each component. By incrementally integrating the corresponding components, we demonstrate their individual effects. ~\cref{fig:supp_ablation_1} showcases representative visual outcomes from our qualitative experiments.  
A comparison between the second and third columns highlights that cross-modal AdaIN significantly improves text alignment while preserving style similarity. Furthermore, as shown in the green apple example, introducing the Teacher Model not only enhances layout stability but also resolves remaining artifacts, ensuring spatial consistency across different styles. 

\begin{figure*}
    \centering
    \includegraphics[width=1.0\linewidth]{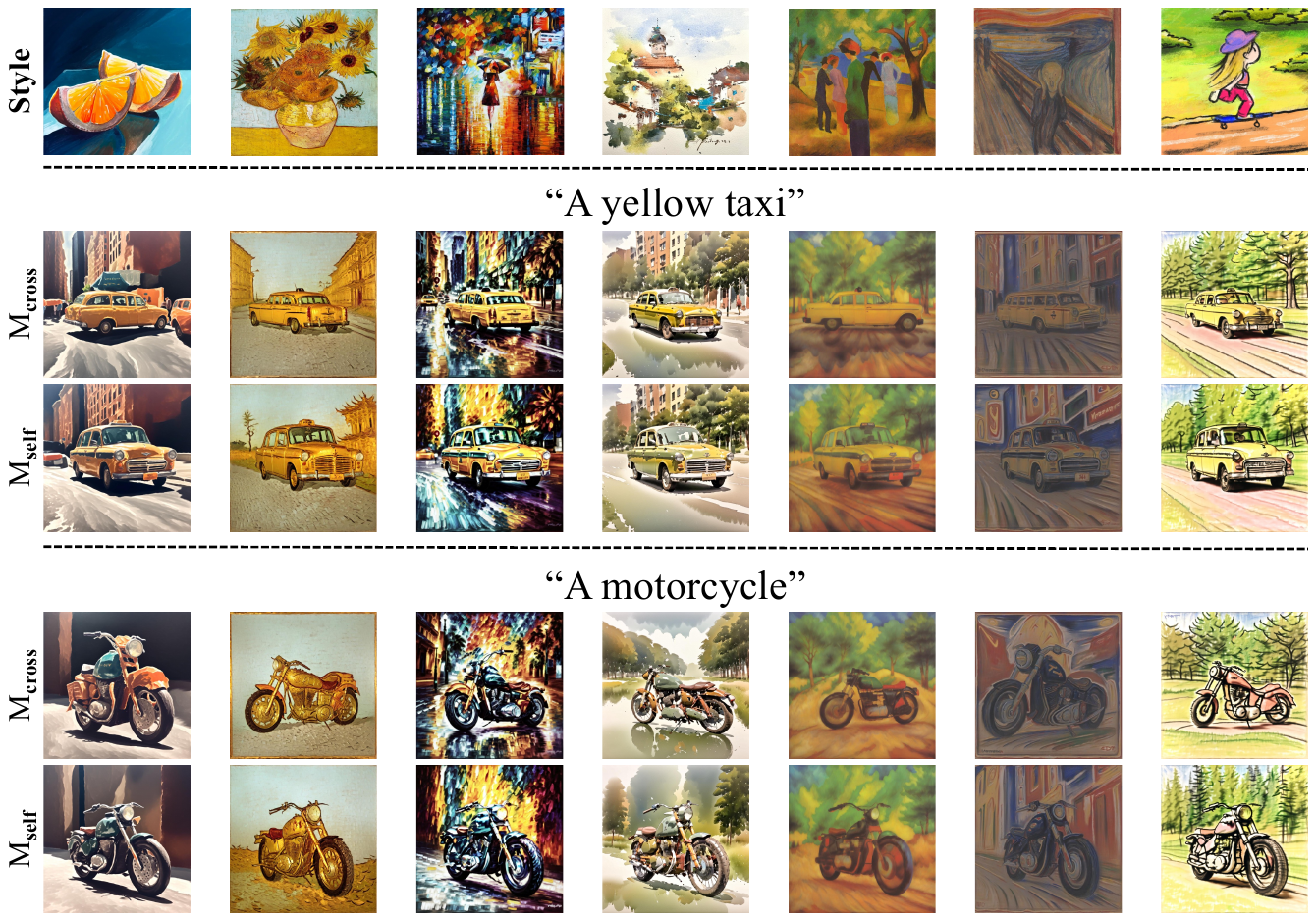}
    \vspace{-7mm}
    \caption{Implementation of the Teacher Model: Comparison of substituting the Self-Attention Map and Cross-Attention Map. The results demonstrate that replacing the Self-Attention Map achieves layout stability and consistency across different styles of images.}
    \label{fig:supp_ablation_attnMap}
    \vspace{-5mm}
\end{figure*}

\noindent \textbf{Self-Attention map and layout stability.} In the UNet of Stable Diffusion~\cite{rombach2022high, podell2023sdxl}, Cross-Attention~\cite{vaswani2017attention} primarily aligns the prompt with the generated image, determining how textual input influences the overall style and content. Self-Attention~\cite{vaswani2017attention}, on the other hand, focuses on the internal coherence of the image, maintaining spatial relationships and structural consistency. As shown in ~\cref{fig:supp_ablation_attnMap}, swapping the Self-Attention Map ensures layout stability and consistency across different styles of images, whereas replacing the Cross-Attention Map fails to achieve this effect, resulting in noticeable differences in the main layout under varying styles. 
All experiments were conducted by adding the Teacher Model to the baseline CSGO framework. To objectively evaluate the impact of the Teacher Model, cross-modal AdaIN was not used in these experiments, isolating the Teacher Model's contribution to layout stability.

\begin{figure*}
    \centering
    \includegraphics[width=1.0\linewidth]{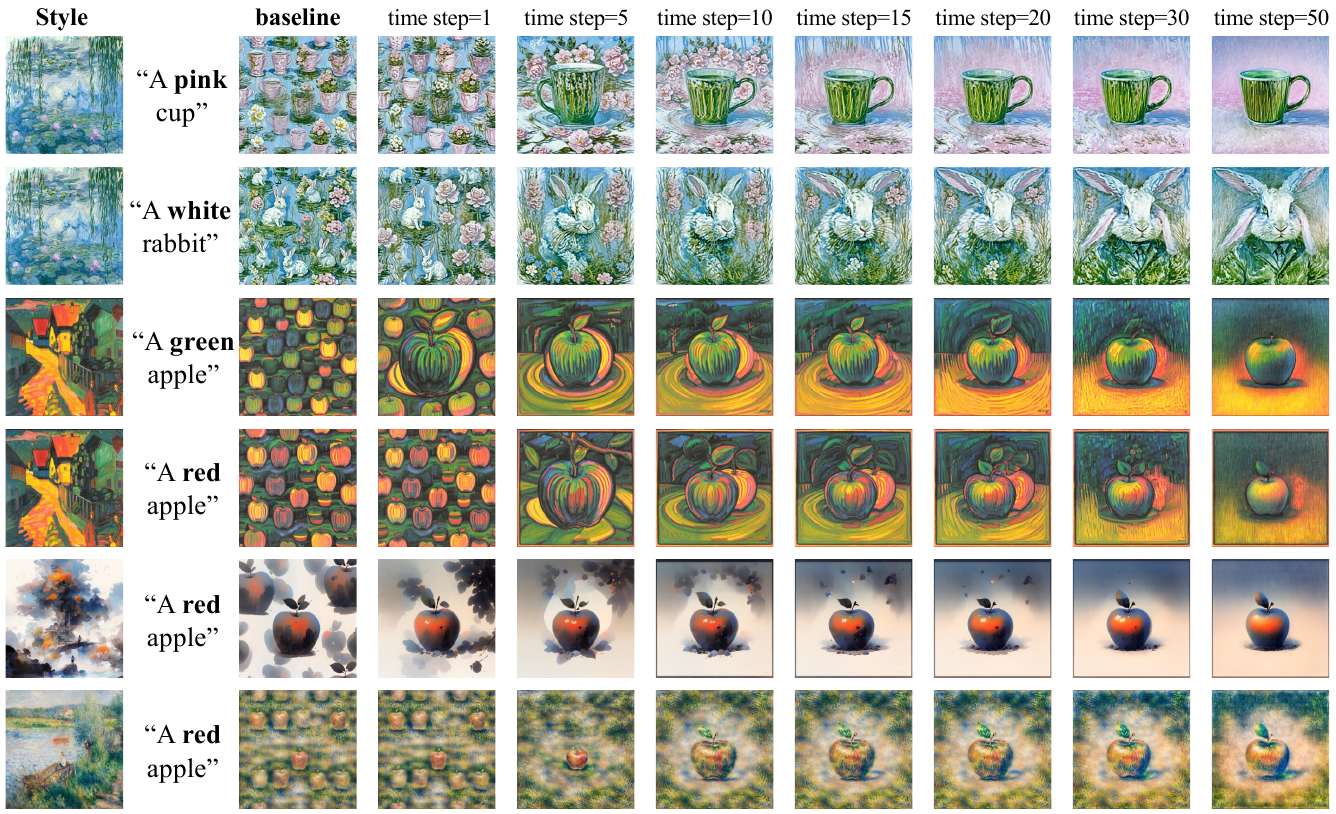}
    \caption{Impact of Teacher Model on Style Image Generation. The term ``timestep'' refers to the number of denoising steps during which the Teacher Model is involved. Notably, these experiments were conducted without incorporating cross-modal AdaIN to isolate and evaluate the specific impact of the Teacher Model on the generated results.}
    \label{fig:supp_ablation_teacher_time}
\end{figure*}

\noindent \textbf{Choice of Teacher Model participation timestep.}
To evaluate the impact of the Teacher Model’s participation timestep on the final generation results, we conducted experiments analyzing its effect. The Teacher Model is designed to ensure layout stability while avoiding artifacts, such as checkerboard patterns. To objectively evaluate the impact of the Teacher Model, cross-modal AdaIN was not used in these experiments.
As shown in ~\cref{fig:supp_ablation_teacher_time}, the term ``timestep'' refers to the number of denoising steps during which the Teacher Model is active. The results demonstrate that insufficient participation (short timesteps) fails to resolve layout issues, while prolonged involvement (long timesteps) negatively affects the final style fidelity. Rows 3 and 4 illustrate that even small changes in the timestep significantly influence the results, while Rows 5 and 6 show that the optimal timestep can vary across different styles. Based on these findings, a timestep between 10 and 20 strikes a reasonable balance between layout stability and style preservation.

\begin{figure*}
    \centering
    \includegraphics[width=1.0\linewidth]{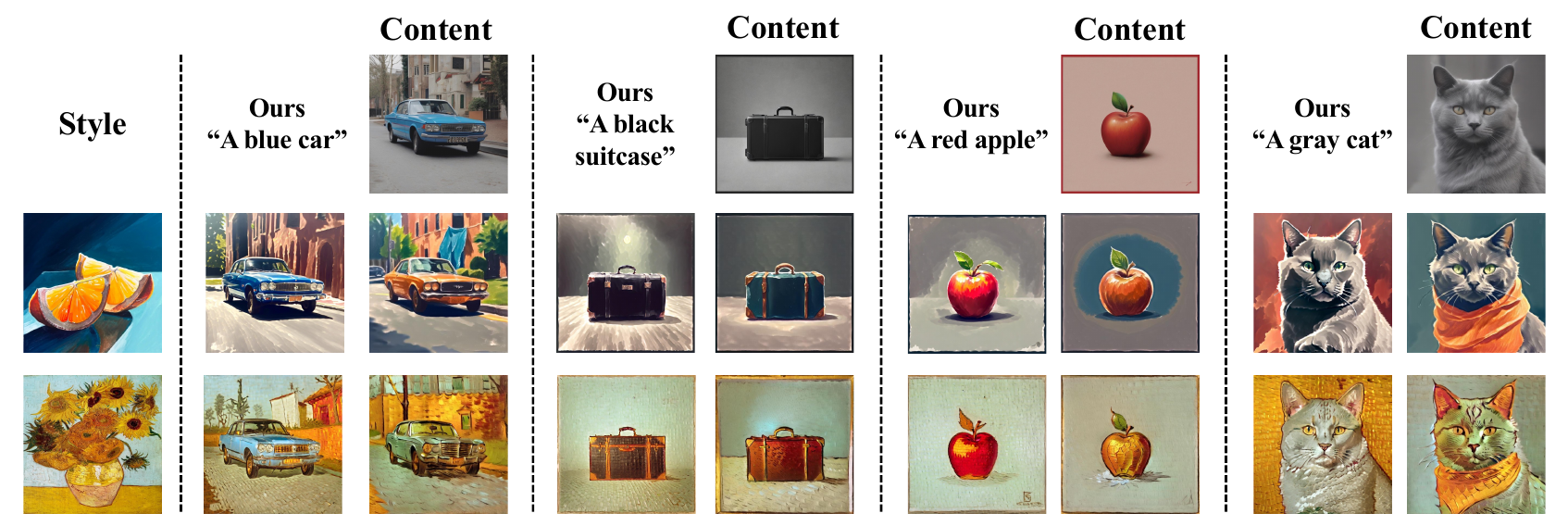}
    \caption{Compared to the image-based style transfer(I2I) provided by CSGO~\cite{xing2024csgo}, We ensured the use of the same initial noise for both our method and the generation of the content image for I2I. It can be observed that the results obtained using the Teacher Model differ significantly from those of I2I, as I2I fails to preserve the color information of the original image.}
    \label{fig:supp_i2i_compare}
    \vspace{-5mm}
\end{figure*}

\noindent \textbf{Compare with image-based style transfer(I2I).} Although our method utilizes the Self-Attention Map provided by the Teacher Model, this does not equate to I2I. As shown in ~\cref{fig:supp_i2i_compare}, the I2I approach provided by CSGO~\cite{xing2024csgo} fails to preserve the color information of the content image effectively. In contrast, our method can more accurately adhere to the prompt's description. To ensure fairness, the noise used in our method is identical to that used in generating the content image.

\section{Additional Comparisons}
\label{sec:Add_Compare}
Qualitative experiments are conducted to visually demonstrate the strengths of our method, particularly in capturing style details and ensuring alignment with the given textual descriptions. This allows for a more intuitive comparison with state-of-the-art methods, showcasing the superior performance of our approach in real-world scenarios. 
We provided additional qualitative comparisons between our method and state-of-the-art approaches to better illustrate the strengths and weaknesses of each method. 

\begin{figure*}
    \centering
    \includegraphics[width=1.0\linewidth]{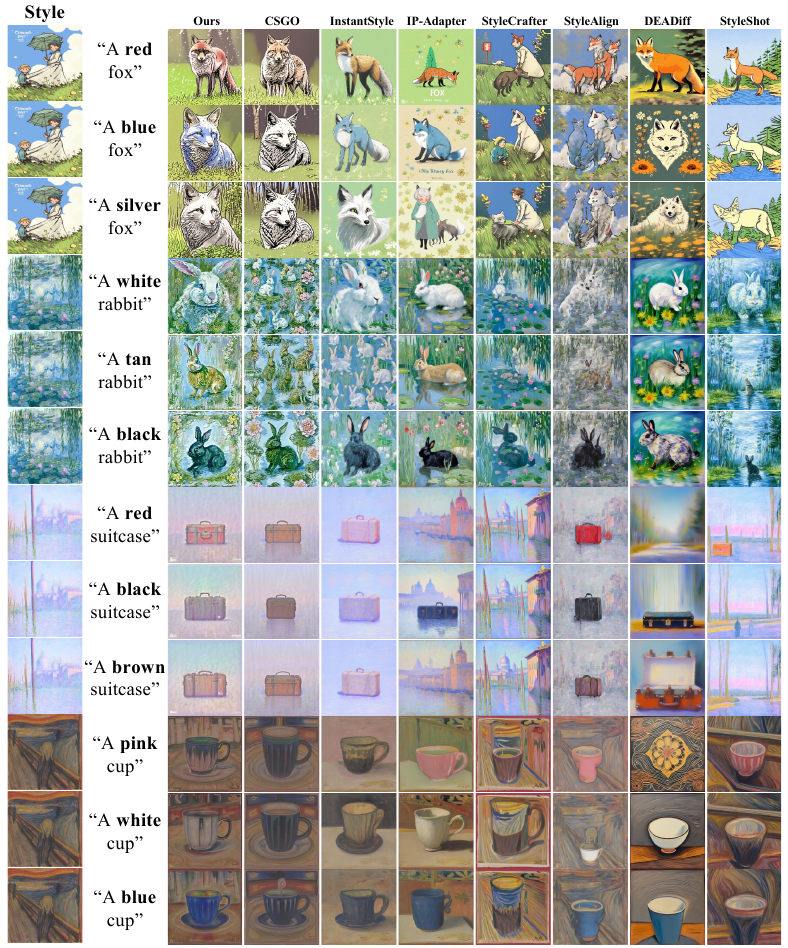}
    \vspace{-7mm}
    \caption{Qualitative comparison with state-of-the-art methods. Our approach effectively preserves image style while accurately adhering to text prompts for generation.}
    \label{fig:supp_quality_1}
    \vspace{-5mm}
\end{figure*}

In ~\cref{fig:supp_quality_1}, our method outperforms others in both overall style similarity and the ability to capture fine details, such as textures. Additionally, it achieves the highest accuracy in aligning with the prompt descriptions.
For methods based on the Stable Diffusion XL~\cite{podell2023sdxl}, approaches like CSGO~\cite{xing2024csgo} and InstantStyle~\cite{wang2024instantstyle} exhibit noticeable style overfitting, while IP-Adapter~\cite{ye2023ip} and StyleCrafter~\cite{liu2023stylecrafter} tend to suffer from content leakage. Meanwhile, StyleAlign~\cite{wu2021stylealign} produces results of relatively lower quality.
For methods based on the Stable Diffusion 1.5, DEADiff~\cite{qi2024deadiff} struggles with accurately capturing the style, and although StyleShot~\cite{gao2024styleshot} performs reasonably well in capturing style, it still encounters issues such as content leakage.
Content leakage can indeed be seen as a form of overfitting to the style reference, where the model overly relies on the style image, causing elements of the style reference to dominate or intrude on the content representation. This highlights a lack of proper disentanglement between style and content in such cases. 

\begin{figure*}
    \centering
    \includegraphics[width=1.0\linewidth]{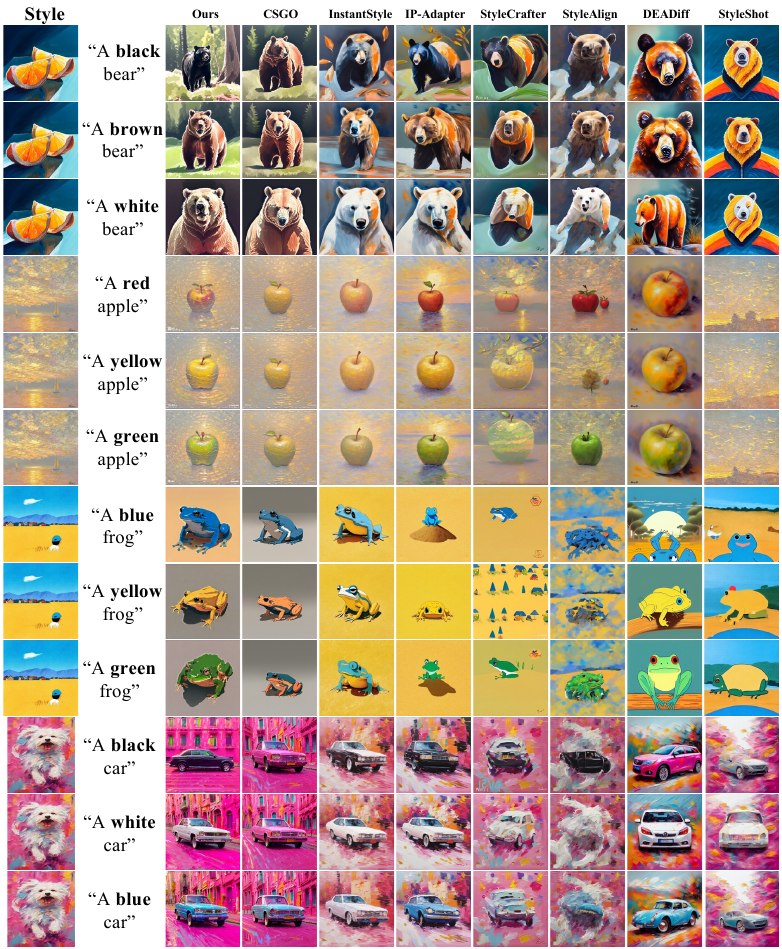}
    \vspace{-7mm}
    \caption{Qualitative comparison with state-of-the-art methods. Our approach effectively preserves image style while accurately adhering to text prompts for generation.}
    \label{fig:supp_quality_2}
    \vspace{-5mm}
\end{figure*}

A more nuanced form of style overfitting, as discussed in this paper, arises when text-driven style transfer methods struggle to adapt to nuanced variations in prompt details, such as changes in color. The challenge lies in whether these methods can accurately align with the evolving prompt descriptions while preserving the integrity of the style. 
This aspect is further validated in ~\cref{fig:supp_quality_2}. Methods such as CSGO~\cite{xing2024csgo}, InstantStyle~\cite{wang2024instantstyle}, and StyleShot~\cite{gao2024styleshot} struggle to differentiate the color specifications described in the prompt. Additionally, IP-Adapter~\cite{ye2023ip} and DEADiff~\cite{qi2024deadiff} face challenges with style dissimilarity, while StyleCrafter~\cite{liu2023stylecrafter} demonstrates some bias toward the structure of the style reference, particularly evident in the ``car'' example.

\begin{figure*}
    \centering
    \includegraphics[width=1.0\linewidth]{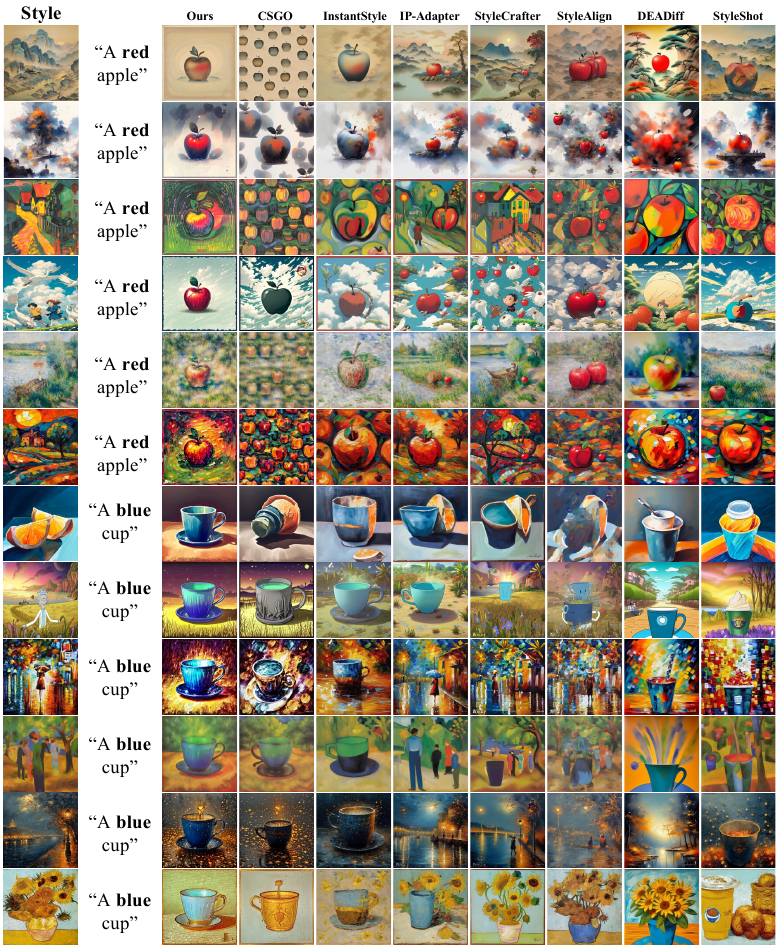}
    \vspace{-7mm}
    \caption{Qualitative comparison with state-of-the-art methods. Our approach effectively maintain layout consistency across different styles under the same prompt.}
    \label{fig:supp_quality_3}
    \vspace{-5mm}
\end{figure*}

In the main paper, we also focus on the issue of layout stability. Through extensive experiments, as shown in ~\cref{fig:supp_quality_3}, we demonstrate that our method can effectively ensure layout stability. CSGO~\cite{xing2024csgo} frequently exhibits artifacts such as checkerboard patterns, while other methods also encounter issues with layout instability. Notably, content leakage appears to be closely related to layout disruptions. This can be validated from the experimental results of StyleCrafter~\cite{liu2023stylecrafter} and IP-Adapter~\cite{ye2023ip}. Although ``A red apple'' is reflected in the final generated output, the image contains too many unrelated elements from the style reference, making it appear overly cluttered.

\section{More results from our study}
\label{sec:more_result}
\begin{figure*}
    \centering
    \includegraphics[width=1.0\linewidth]{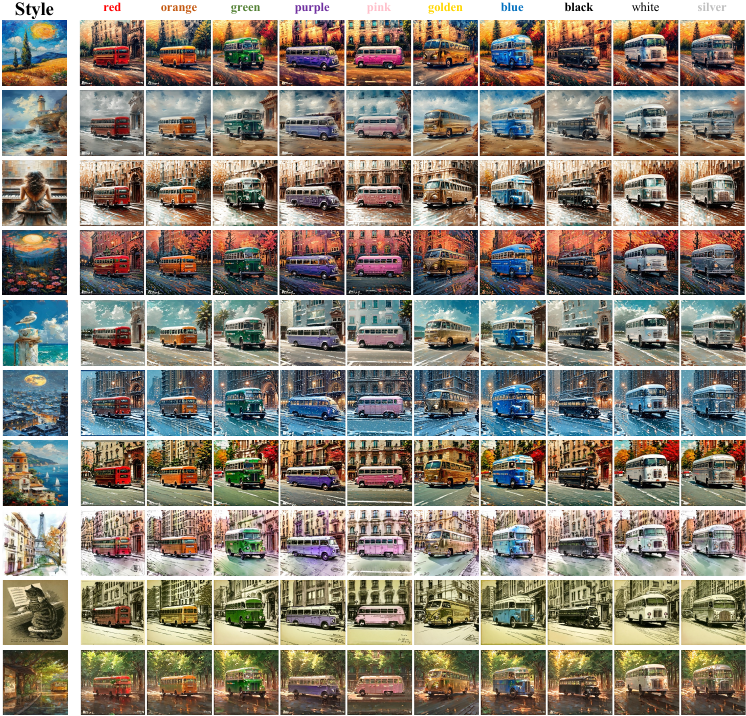}
    \vspace{-7mm}
    \caption{More results of our text-driven style transfer model. Given a style reference image, our method effectively reduces style overfitting, generating images that faithfully align with the text prompt while maintaining consistent layout structure across varying styles. Illustration of the prompt format used: ``A [color] bus''.}
    \label{fig:supp_fig_ours1}
    \vspace{-5mm}
\end{figure*}

\begin{figure*}
    \centering
    \includegraphics[width=1.0\linewidth]{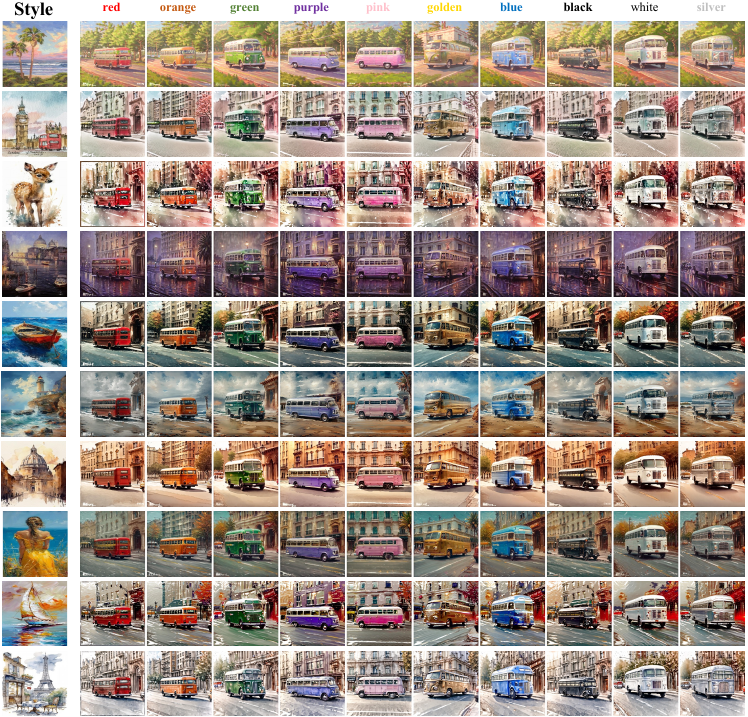}
    \vspace{-7mm}
    \caption{More results of our text-driven style transfer model. Given a style reference image, our method effectively reduces style overfitting, generating images that faithfully align with the text prompt while maintaining consistent layout structure across varying styles. Illustration of the prompt format used: ``A [color] bus''.}
    \label{fig:supp_fig_ours2}
    \vspace{-5mm}
\end{figure*}

In ~\cref{fig:supp_fig_ours1}, ~\cref{fig:supp_fig_ours2}, and ~\cref{fig:supp_scfg_more} we provide additional visualization results showcasing the effectiveness and versatility of our method. We have selected a variety of style categories and different color schemes to highlight the alignment effects for text descriptions. Moreover, we achieve excellent layout stability even when using the same prompt.

\section{Integration with Other Methods}
\label{sec:integration}
CSGO~\cite{xing2024csgo} has been recognized as one of the most effective and state-of-the-art methods for style transfer, which is why it was selected as the primary baseline in the main paper. To further evaluate the generalizability and robustness of our approach, we additionally explored its application and performance on other models.
\subsection{Integration with InstantStyle~\cite{wang2024instantstyle}}
\begin{figure*}
    \centering
    \includegraphics[width=1.0\linewidth]{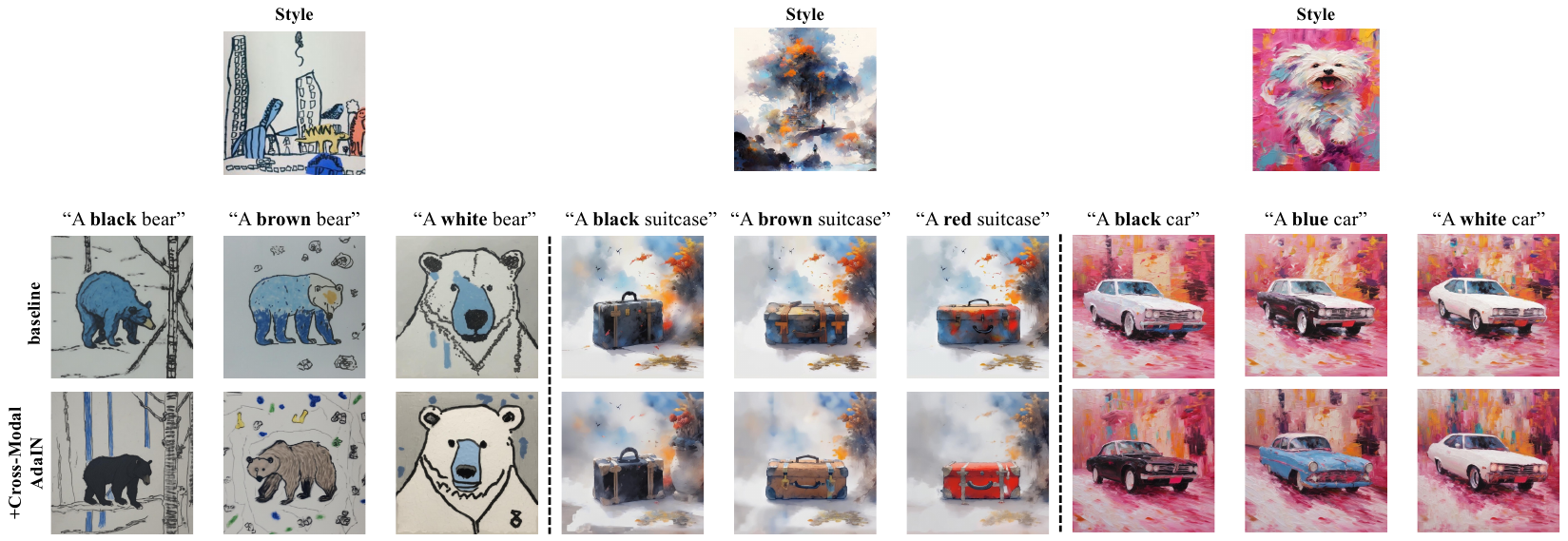}
    \vspace{-8mm}
    \caption{Qualitative results of using cross-modal AdaIN in InstantStyle~\cite{wang2024instantstyle}. The results demonstrate that cross-modal AdaIN effectively prevents style overfitting. The final generated results consistently align with the textual descriptions.}
    \label{fig:supp_instantstyle_abl_1}
    \vspace{-5mm}
\end{figure*}

\noindent \textbf{Cross-Modal AdaIN.}
Since InstantStyle~\cite{wang2024instantstyle} is also an adapter-based architecture, it can similarly integrate cross-modal AdaIN to mitigate style overfitting. The results are shown in ~\cref{fig:supp_instantstyle_abl_1}. Compared to Row 1, Row 2 accurately follows the text description, effectively avoiding errors in the generated output.

\begin{figure*}
    \centering
    \includegraphics[width=1.0\linewidth]{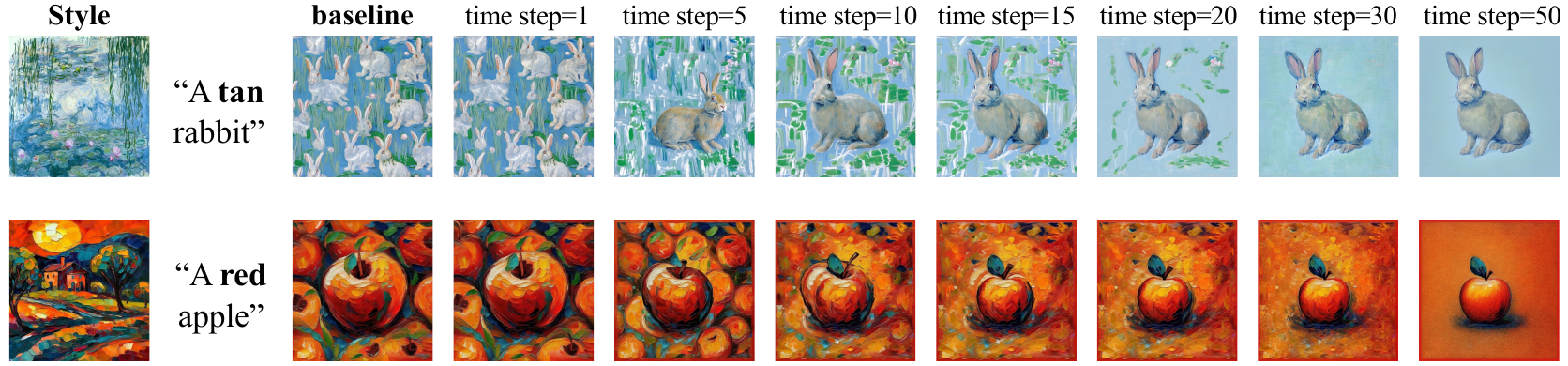}
    \vspace{-8mm}
    \caption{Impact of Teacher Model on InstantStyle~\cite{wang2024instantstyle} Image Generation. The term “timestep” refers to the number of denoising steps during which the Teacher Model is involved. Notably, these experiments were conducted without incorporating cross-modal AdaIN to isolate and evaluate the specific impact of the Teacher Model on the generated results. When the Teacher Model is applied to InstantStyle~\cite{wang2024instantstyle}, it helps prevent the generation of artifacts, such as checkerboard patterns.}
    \label{fig:supp_instantstyle_teacher}
    \vspace{-5mm}
\end{figure*}

\noindent \textbf{Teacher Model.}
InstantStyle~\cite{wang2024instantstyle} also encounters artifacts such as checkerboard patterns. Similar to the previous approach, we investigated the impact of the Teacher Model's involvement at different timesteps on the results, as shown in ~\cref{fig:supp_instantstyle_teacher}. Upon observation, we reached a similar conclusion: if the Teacher Model participates for too many timesteps, it can lead to style loss.

\subsection{Integration with StyleCrafter~\cite{liu2023stylecrafter}}
\begin{figure*}
    \centering
    \includegraphics[width=1.0\linewidth]{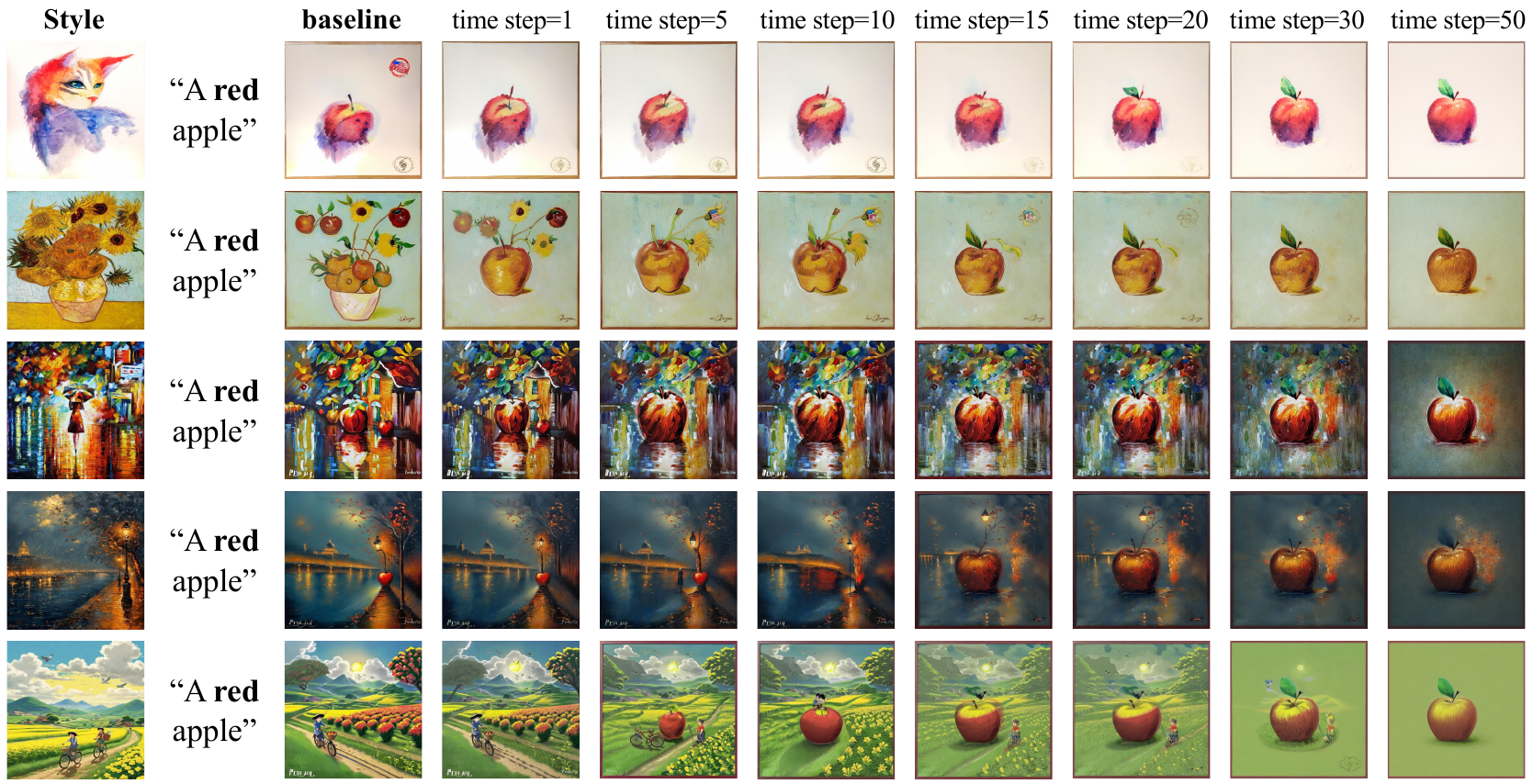}
    \vspace{-8mm}
    \caption{Impact of Teacher Model on StyleCrafter~\cite{liu2023stylecrafter} Image Generation. The term “timestep” refers to the number of denoising steps during which the Teacher Model is involved. Notably, these experiments were conducted without incorporating cross-modal AdaIN to isolate and evaluate the specific impact of the Teacher Model on the generated results. In addition to ensuring layout stability, the Teacher Model also effectively reduces the occurrence of content leakage when applied to StyleCrafter~\cite{liu2023stylecrafter}.}
    \label{fig:supp_stylecrafter_teacher}
    \vspace{-3mm}
\end{figure*}

\noindent \textbf{Teacher Model.} A notable issue in StyleCrafter~\cite{liu2023stylecrafter} is content leakage, where unrelated content elements from the style image appear in the generated results, ultimately affecting the final output. This phenomenon can lead to generated images that do not align with the descriptions in the prompt. To address this, we incorporated the Teacher Model into the method. As shown in ~\cref{fig:supp_stylecrafter_teacher}, the inclusion of the Teacher Model significantly mitigates the problem of content leakage, resulting in outputs that maintain stability and consistency across different styles.

